%% file: main.tex
\pdfoutput=1

\documentclass[11pt]{article}
\usepackage[preprint]{acl}

\input{macro}

\title{Babysit A Language Model From Scratch: \\Interactive Language Learning by Trials and Demonstrations}

\author{%
  Ziqiao Ma\thanks{Authors contributed equally to this work.} \\
  University of Michigan\\
  \texttt{marstin@umich.edu}
  \And
  Zekun Wang\footnotemark[1] \\
  University of Michigan\\
  \texttt{zekun@umich.edu}
  \And
  Joyce Chai \\
  University of Michigan\\
  \texttt{chaijy@umich.edu}
}

\begin{document}
\maketitle

\input{sec/00-abstract}
\input{sec/01-introduction}
\input{sec/03-method}

\input{sec/04-experiment}
\input{sec/02-background}

\input{sec/06-conclusion}
\input{sec/07-acknowledge}

\bibliography{references}

\clearpage
\appendix
\input{sec/a1-method}

\input{sec/a2-experiment}
\input{sec/a3-ethics}

\end{document}

%% file: macro.tex
\usepackage{times}
\usepackage{latexsym}

\usepackage[T1]{fontenc}

\usepackage[utf8]{inputenc}

\usepackage{microtype}

\usepackage{inconsolata}

\usepackage{graphicx}

\usepackage{hyperref}       
\usepackage{url}            
\usepackage{booktabs}       
\usepackage{amsfonts}       
\usepackage{nicefrac}       
\usepackage{microtype}      
\usepackage{xcolor}         

\usepackage{rotating}
\usepackage{tcolorbox}
\usepackage{algorithm}
\usepackage{algpseudocode}
\usepackage{amsmath}
\usepackage{algorithmicx}
\usepackage{booktabs}
\usepackage{multirow}
\usepackage{multicol}
\usepackage{amsfonts}
\usepackage{cleveref}

\usepackage{wrapfig,lipsum}

\usepackage{subcaption}

\usepackage[shortlabels]{enumitem}
\setlist[itemize]{leftmargin=*}
\setlist[enumerate]{leftmargin=*}
\usepackage[normalem]{ulem}

\definecolor{commentgrey}{gray}{0.7}
\newcommand{\greycomment}[1]{\textcolor{commentgrey}{\textit{#1}}}


%% file: sec/00-abstract.tex
\begin{abstract}

Humans are efficient language learners and inherently social creatures. 
Our language development is largely shaped by our social interactions, for example, the demonstration and feedback from caregivers. 
Contrary to human language learning, recent advancements in large language models have primarily adopted a non-interactive training paradigm, and refined pre-trained models through feedback afterward.
In this work, we explore how corrective feedback from interactions influences neural language acquisition from scratch through systematically controlled experiments, assessing whether it contributes to word learning efficiency in language models. 
We introduce a trial-and-demonstration (TnD) learning framework that incorporates three distinct components: student trials, teacher demonstrations, and a reward conditioned on language competence at various developmental stages. 
Our experiments reveal that the TnD approach accelerates word acquisition for student models of equal and smaller numbers of parameters, and we highlight the significance of both trials and demonstrations. 
We further show that the teacher's choices of words influence students' word-specific learning efficiency, and a practice-makes-perfect effect is evident by a strong correlation between the frequency of words in trials and their respective learning curves. 
Our findings suggest that interactive language learning, with teacher demonstrations and active trials, can facilitate efficient word learning in language models.

\end{abstract}

%% file: sec/01-introduction.tex
\section{Introduction}

Humans are social beings and we learn language from interactions~\citep{vygotsky1934thought,bruner1985child,palincsar1986role,kuhl2004early,tomasello2005constructing}.
Long before children's linguistic skills are mature, they could engage in early forms of conversational exchange with others~\citep{halliday1975learning,clark2018conversation}. 
A critical component of social interactions that language grounds to is the \textit{feedback} provided by the caregivers~\citep{warlaumont2014social}.
This, for example, includes the communicative feedback~\citep{bates1975acquisition,nikolaus2023communicative} that highlights the success and failure of communication, and the corrective feedback~\citep{farrar1992negative,chouinard2003adult,saxton2005negative,hiller2016corrective} that is more direct and emphasizes the responses from caregivers, which offer corrections to possible errors in children's speech, encompassing variants like negative evidence, reformulations, or recasts.
Unlike human learners who acquire language skills through feedback during interactions, most language models differ in terms of their inductive biases and data sources~\citep{warstadt2022artificial}.
These models typically learn from massive text corpora using cross-entropy loss for self-supervised learning.

\input{sec/input/framework}

Recently, several lines of cognitively motivated language modeling research have looked into the learnability and learning efficiency of language~\cite{portelance2020predicting,chang2022word,evanson2023language}.
By incorporating non-linguistic inputs such as multimodal stimuli~\cite{shi2019visually,ma2023world,portelance2023learning} and/or communicative feedback~\cite{nikolaus2021modeling,zhu2022language,liu2022computational}, recent studies have explored potential mechanisms that contribute to efficient language learning in (vision-)language models.
Through controlled ablation studies~\cite{warstadt2022artificial}, these models can serve as proof of concept to verify mechanisms that are practically effective for machines, and generate hypotheses that are possible for cognitive learners~\citep{portelance2022neural,portelance2023roles}.
In a similar spirit, we seek to investigate the role of explicit corrective feedback in neural language learning through controlled computational experiments.
Rather than making direct comparisons to human learning, our goal is to examine if student trials and teacher demonstrations promote efficient word learning in language model training, and if so, which words benefit the most.

We introduce Trial-and-Demonstration (TnD), an interactive learning framework that incorporates corrective feedback with three components: student model \textit{trials}, teacher model \textit{demonstrations}, and a \textit{reward} conditioned on the training trajectory of the model (Figure~\ref{fig:tnd}). 
Our experiments reveal that the TnD approach accelerates word acquisition, highlighting the significance of both trials and demonstrations. 
From the teacher's perspective, their word choices affect students' word-specific learning efficiency. 
From the student's side, the word frequency in trials closely aligns with the learning curve, supporting the idea that learning by language production accelerates word proficiency. 
Our findings highlight that trials and demonstrations can facilitate word learning in language models, and further, suggest an efficient alternative to building language models interactively.


%% file: sec/input/framework.tex
\begin{figure*}[!t]
    \centering
    \includegraphics[width=\linewidth]{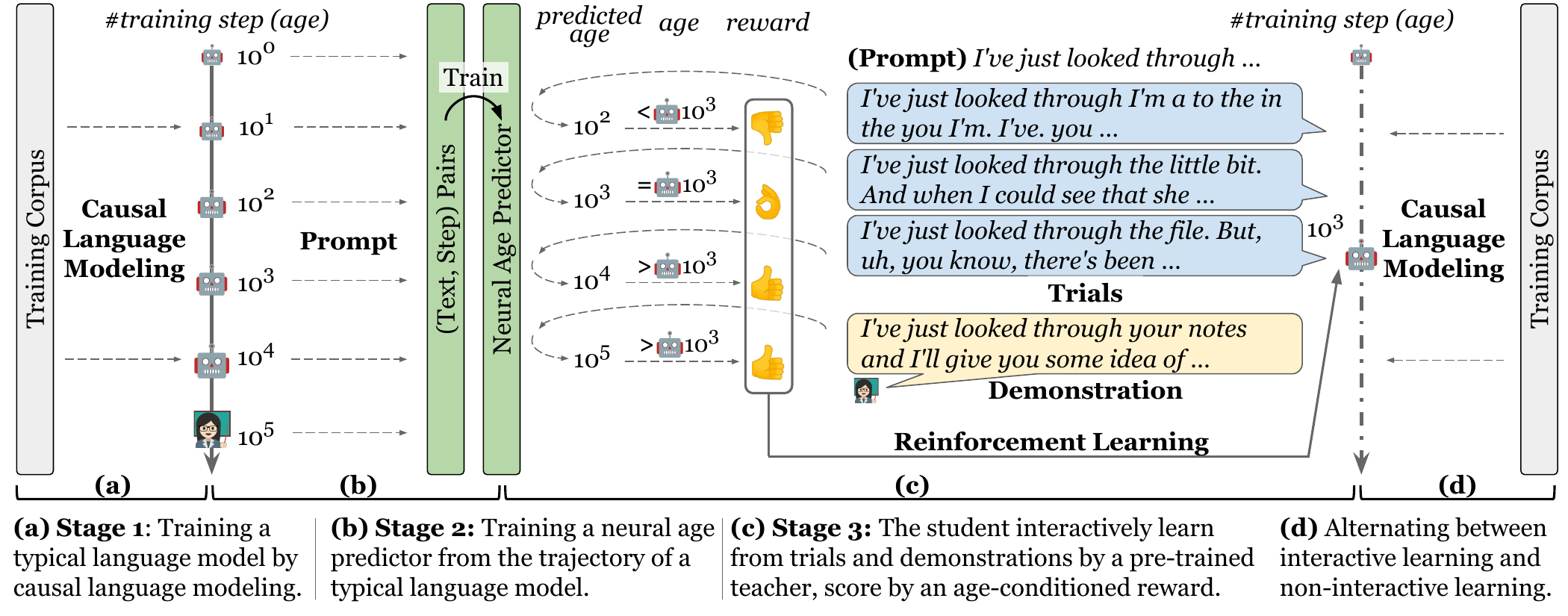}
    \vspace*{-20pt}
    \caption{The learning by trial-and-demonstration (TnD) framework. In stage 1, we start by training a language model with the causal language modeling objective. In stage 2, we prompt the models along the learning trajectory for (text, step) pairs and train a neural age predictor to predict the training step given a text. In stage 3, we use the final model in stage 1 as the teacher model. In an interactive step, the student model is prompted to complete a trial, and the teacher model is prompted to provide a demonstration. The trials and demonstrations are scored by an age-conditioned reward function (Eq.~\ref{eq:reward}), and the student model updates the policy with reinforcement learning. The student alternates between interactive and non-interactive steps. \vspace*{-15pt}}
    \label{fig:tnd}
\end{figure*}


%% file: sec/03-method.tex
\vspace{-5pt}
\section{Interactive Language Learning by Trials-and-Demonstrations (TnD)}
\label{sec:tnd}
\vspace*{-5pt}

Modeling corrective feedback in computational language learning presents significant challenges, as recruiting human subjects to supervise the development of a language model from the ground up over numerous iterations is impractical. 
Consequently, we present a human-free Trial-and-Demonstration (TnD) learning framework that streamlines the process (Figure~\ref{fig:tnd}).
In a scenario of corrective feedback, the student model engages in production-based learning: to produce an initial utterance, followed by the teacher model generating its version of the text as a demonstration. 
For the student model to recognize the teacher's response as preferable and to facilitate learning, these language outputs are evaluated by a reward function, which is based on the competence of the student's language use that is expected for its developmental stage (i.e., training steps).
The TnD framework thus includes three components: a student model's \textit{trials}, a teacher model's \textit{demonstrations}, and \textit{rewards}. 
This framework allows us to incorporate massive corpora to study modern generative language models, offering a general and unrestricted approach to simulate interactive language learning with corrective feedback on massive corpora.

\vspace*{-5pt}
\subsection{The student model and trials}

We employ randomly initialized GPT-2~\citep{radford2019language} as the student model for our investigation into language acquisition, leveraging its causal language modeling (CLM) objective and inherent generative capabilities for production-based learning.
To encourage the student model to attempt text production, it is essential to provide an appropriate context.
In each trial, we prompt the student with the first 5 tokens from a natural sentence, asking it to generate the continuation as a \textit{trial}.

\vspace*{-5pt}
\subsection{The teacher model and demonstrations}

Inspired by recent work~\citep{bai2022constitutional,lee2023rlaif,saha2023can}, we utilize pre-trained language models as proxies for human language teachers.
Employing language models as ``caregivers'' for language models offers two advantages.
Firstly, it eliminates the need for recruiting human participants to engage with a language model across thousands of iterations.
Secondly, we can consistently control the behavior of the teacher model across experiments, and adjust the teacher's language behaviors by modifying its decoding strategies for language generation.
The process of developing a teacher model is identical to the typical language model pre-training, as shown in Figure~\ref{fig:tnd}(a).
We adopted the same GPT-2 architecture and pre-trained the model with the CLM objective for 100k steps, with all hyper-parameters following the default setup.
To generate a natural language \textit{demonstration} for the student's \textit{trial}, we prompt the pre-trained teacher model with the same 5 tokens used for the student model, thereby obtaining the teacher's completion of the sentence.

\vspace*{-2pt}
\subsection{The reward and reward model}
\label{sec:reward}

\input{sec/input/reward}

Defining an effective reward in our context is challenging due to the absence of communication games and the lack of access to large-scale human preference annotations.
Heuristic reward metrics do not consider the developmental trajectory of language models, which is critical for simulating language acquisition. 
It's akin to human development where early words are celebrated as milestones, but prolonged reliance on initial language abilities can become a concern. 
We treat the number of training steps as the neural model's ``age''~\citep{chang2022word}.
A language model that generates fluent text at 500 steps, which typically emerges around 5,000 steps, should be rewarded for its accelerated learning. 
Conversely, if the language production quality in the student remains the same at 50,000 steps, it should be penalized.
This process is illustrated in Figure~\ref{fig:tnd}(c).

To train a neural age predictor, we utilize the developmental trajectory of the teacher model by saving over 100 checkpoints at various training steps. 
For each checkpoint, we select 25,000 contexts (each consisting of 5 tokens) from the test set and prompt the model to generate continuations for each context. This process generates a dataset comprising over 2.5 million (text, step) pairs. 
We then use this dataset to fine-tune a LLaMA-2-7B language model~\citep{touvron2023llama}, incorporating a 1-layer linear head for regression. 
At step $n$, the student model $\pi_\theta$ parameterized by $\theta$ produces a sentence $\mathcal{S}$ of $l$ tokens $t_1,\cdots,t_l$.
We use the neural age predictor $R_{\phi}$ to estimate the logarithm of the expected training step $\hat{n}$ where this sentence typically emerges. Hence, the age-conditioned reward $r(\mathcal{S},n)$ is given by:
\vspace*{-5pt}
\begin{equation}
\label{eq:reward}
    r := \log ( \hat{n} / n ) = R_{\phi}(\mathcal{S}) - \log n
\end{equation}
\vspace*{-5pt}

\input{sec/input/algorithm}

\subsection{Alternating interactive and non-interactive language learning}

As shown in Figure~\ref{fig:tnd}(c-d), our TnD framework alternates between two forms of language learning: 
(1) interactive learning, in which corrective feedback is taken through reinforcement learning, utilizing rewards derived from both the student's trials and the teacher's demonstrations, and 
(2) non-interactive learning, which emulates the natural language exposure experienced by learners and is facilitated through causal language modeling as adopted by generative language models.

\vspace*{-5pt}
\subsubsection{Interactive language learning setup.}
Reinforcement learning has been applied to language models, such as the success of games~\citep{narasimhan2015language,he2016deep}, heuristic scores~\citep{ranzato2016sequence,nikolaus2021modeling}, and models of human preferences~\citep{ziegler2019fine,ouyang2022training}.
The key is to view the language production in language models as actions within a vocabulary-defined action space.
Formally, the model \(\pi_\theta\) is given a context of 5 tokens (i.e., the initial state \(s_0=\{t_1,\cdots,t_5\}\)) to produce the next token (i.e., the next action \(a_1=t_6\)).
It lands in the next state \(s_1=\{t_1\cdots t_6\}\), and this process repeats until the sentence concludes.
Following this, rewards can be computed according to Eq.~\ref{eq:reward} for the student's trials and the teacher's demonstrations.
The goal is to maximize the expected return (i.e., the expected cumulative future reward) following \(\pi_{\theta}\) along the interaction, with the trials and demonstrations both in the training batch. 
Inspired and taking the computational infrastructure in recent advances in reinforcement learning from human feedback (RLHF)~\citep{ouyang2022training}, we use Proximal Policy Optimization (PPO)~\cite{schulman2017proximal} algorithm with a clipped surrogate objective $\mathcal{L}^{\textrm{ppo}}_\theta$, which is the primary variant in modern large language models.\footnote{We use the \href{https://github.com/huggingface/trl/}{TRL} library to train language models with reinforcement learning. Our code is available \href{https://github.com/sled-group/TnD}{here}.}
We applied two modifications in the implementation of the RLHF algorithm, which include the involvement of demonstration in policy update, and the removal of KL-divergence. 
We refer to Appendix~\ref{sec:rl-app} for mathematical details.

\vspace*{-5pt}
\paragraph{Demonstration in policy update.}
The original PPO algorithm only learns from the agent's own trials, i.e., the sentences it generated before.
We expand the training batch to add teacher's demonstrations: At each step, we sample the sentences generated by the latest student model and also collect those from the teacher model with the same prompts. 
Subsequently, we regard both of them as the training batch, and apply the policy update to improve the student model. 
Intuitively, our goal is to encourage the student model to imitate and repeat the teacher's demonstration during training.
As there is a reward disparity between the student's trial and the teacher's demonstration, we motivate the student model to learn a better language generation policy.

\vspace*{-5pt}
\paragraph{Removal of KL-divergence objective.}
The loss function in the conventional RLHF algorithm involves a KL-divergence term between the updated student model and a reference model, which is usually a fine-tuned language model or the initial student model before gradient updates.
The goal is to penalize the learned policy that largely deviates from the referenced policy. 
Different from the conventional approach, this penalty is not preferred in the TnD setting. 
This change encourages the student model to explore unfamiliar words during the training, which enables relatively significant updates, as well as eliminates biases from over-adherence to a reference model.

\vspace*{-2pt}
\subsubsection{Non-interactive language learning setup.}
While our interactive language learning replicates active engagement with language through corrective feedback, it's also crucial to simulate human's passive exposure to language, as emphasized in prior psychological~\citep{smith2008infants} and computational~\citep{nikolaus2021modeling} studies.
We implement this with the causal language modeling objective, which is adopted by most generative language models.
Consider a sentence \(t_1,\cdots,t_l\) in the corpus with \(l\) tokens.
The causal language modeling objective is to predict the next token \(t_{i+1}\) given the previous context \(t_{\leq i}\) by minimizing:

\vspace*{-10pt}
\begin{equation}
\label{eq:clm}
    \mathcal{L}^{\text{clm}}_\theta = -\sum_{i=1}^{l-1} \log P(t_{i+1} | t_{\leq i}; \theta)
\end{equation}
\vspace*{-20pt}

\subsubsection{Alternating interactive and non-interactive learning.}
Following the setup of~\citet{nikolaus2021modeling}, we adopt an alternating learning schedule over interactive and non-interactive language learning, i.e., perform $c$ steps of causal language modeling, followed by $r$ steps of reinforcement learning, in a continuous cycle (Algorithm~\ref{alg:tnd}).
We select $c=3$ and $r=1$, respectively, leading to 1 PPO update following every 3 CLM updates. 
We justify this design choice and present experiments on other hyper-parameters in Appendix~\ref{sec:parameters}.

%% file: sec/input/reward.tex

\begin{figure}[!t]
    \centering
    \begin{subfigure}[t]{.23\textwidth}
        \centering
        \includegraphics[width=1.05\linewidth]{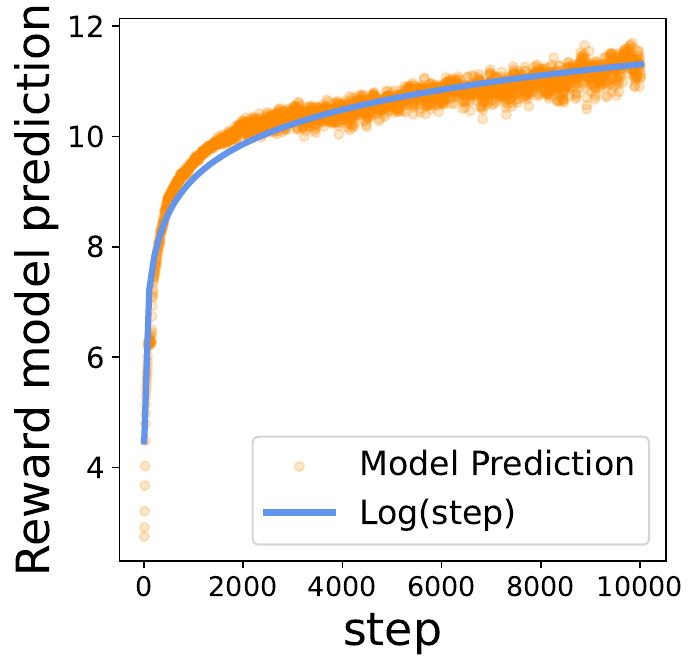}
        \vspace*{-20pt}
        \caption{BookCorpus dataset.}
        \label{fig:reward-bkps}
    \end{subfigure}
    ~
    \begin{subfigure}[t]{.23\textwidth}
        \centering
        \includegraphics[width=1.05\linewidth]{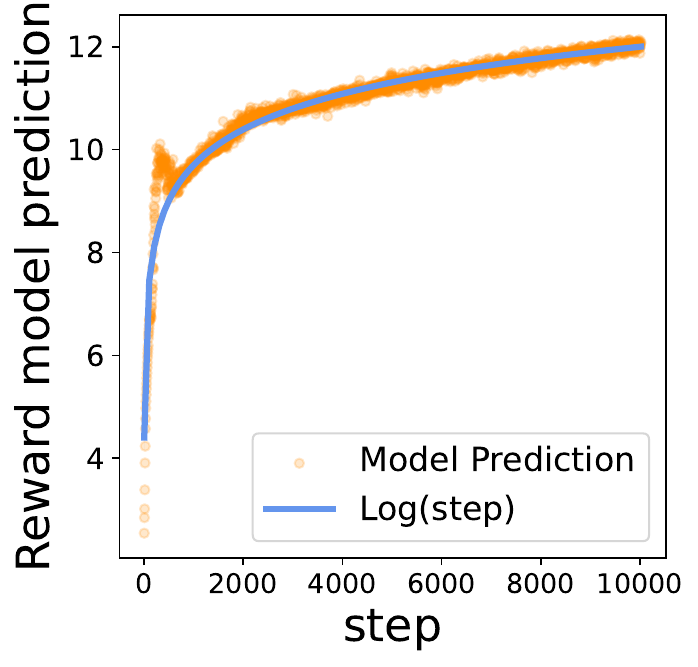}
        \vspace*{-20pt}
        \caption{BabyLM dataset.}  
        \label{fig:reward-babylm}
    \end{subfigure}   
    \vspace*{-8pt}
    \caption{We sample reward model predictions at different steps and compare them to ground truth logarithm. The reward models are satisfactory as the model predicted age/step highly overlaps with the true age/step.\vspace*{-22pt}}
    \label{fig:reward}
\end{figure}

%% file: sec/input/algorithm.tex
\begin{algorithm}[t!]
    \caption{\textsc{Trial-and-Demonstration}}
    \label{alg:tnd}
    \begin{algorithmic}[1]
    \small
        \State \textbf{Input:} student model $\pi_\theta$, teacher model $\hat{\pi}_\varphi$, reward model $R_{\phi}$, alternating schedule $(c,r)$, training corpus $\mathcal{C}$. 
        \For{$n, \mathcal{S}$ \textbf{ in enumerate} $(\mathcal{C})$}
            \State $t_1,\cdots,t_l = \textsc{Tokenize}(\mathcal{S})$
            \If{$n \% (c+r) \leq c$}
                \State \greycomment{// Non-interactive learning}
                \State Gradient descent $\nabla_{\theta}\ \mathcal{L}^{\text{clm}}_\theta([t_1,\cdots,t_l])$ (Eq.~\ref{eq:clm})
            \Else
                \State \greycomment{// Interactive learning}
                \State $\mathcal{S}_{\textrm{trial}} \gets \textsc{Prompt}\big(\pi_\theta, [t_1,t_2,\cdots,t_5]\big)$
                \State $r_{\textrm{trial}} = R_{\phi}(\mathcal{S}_{\textrm{trial}}) - \log n$
                \State $\mathcal{S}_{\textrm{demo}} \gets \textsc{Prompt}\big(\hat{\pi}_\varphi, [t_1,t_2,\cdots,t_5]\big)$
                \State $r_{\textrm{demo}} = R_{\phi}(\mathcal{S}_{\textrm{demo}}) - \log n$
                \State Train batch $B = (S_{\textrm{trial}}, r_{\textrm{trial}}) \cup (S_{\textrm{demo}}, r_{\textrm{demo}})$
                \State Policy update $\nabla_{\theta}\ \mathcal{L}^{\text{ppo}}_\theta(B)$ (See~\ref{sec:rl-app})
            \EndIf
        \EndFor
        \State \textbf{Output:} $\pi_\theta$
    \end{algorithmic} 
\end{algorithm}
\vspace{-20pt}

%% file: sec/04-experiment.tex
\vspace*{-5pt}
\section{Experiment and Evaluations}
\label{sec:exp-eval}

\input{sec/input/surprisal-main}
\input{sec/input/aoa-figure}
\input{sec/input/vocab}

\vspace*{-5pt}
\subsection{Experiment setup}
\label{sec:exp-setup}

\vspace*{-2pt}
\paragraph{Training corpora.}
We repeat our study on two training corpora: 
(1) the BookCorpus~\citep{zhu2015aligning}, which is commonly used for training neural language models; and 
(2) the BabyLM Corpus, a more developmentally plausible dataset provided by the BabyLM Challenge (Round 1)~\citep{warstadt2023findings}.
Notably, the BabyLM dataset has less than 100M words and contains a higher proportion of transcribed speech, e.g., the CHILDES corpus~\citep{macwhinney2000childes}.
For each corpus, we keep 80\% for training and 20\% for evaluation.

\vspace*{-5pt}
\paragraph{Baselines and ablation variants.}
To reliably assess the importance or impact of a mechanism in a language learning system, computational experiments should be conducted under controlled ablation studies~\citep{warstadt2022artificial}.
We describe our baselines below and study other possible setups in Appendix~\ref{sec:baselines}.
\vspace*{-5pt}
\begin{itemize}[leftmargin=*]
    \setlength\itemsep{-0.25em}
    \item The \texttt{CLM} model, which adheres to the original GPT-2 pre-training with only CLM objective;
    \item The \texttt{TnD} model, which implements the trial-and-demonstration framework described in Section~\ref{sec:tnd};
    \item The \texttt{Trial} model, which is the \texttt{TnD} framework with only student trials (no demonstrations);
    \item The \texttt{Demo} model, which is the \texttt{TnD} framework with only teacher demonstrations (no trials);
\end{itemize}
\vspace*{-5pt}
These baselines are designed in a controlled manner to ensure fair ablations.
For each combination of corpus and baseline, the training process is conducted on 5 random seeds for 10k steps.
We discuss more about other possible baseline setups and the hyper-parameters in Appendix~\ref{sec:experiment-appendix}.

\vspace*{-2pt}
\subsection{Evaluation}

\vspace*{-3pt}
\paragraph{Testing vocabulary.}
We specify two sets of vocabulary for evaluation. 
\vspace*{-5pt}
\begin{itemize}[leftmargin=*]
    \setlength\itemsep{-0.25em}
    \item The \texttt{CMN} set, consisting of common words that appear frequently in both corpora, covering a wide range of words and parts of speech.
    \item The \texttt{CDI} set, consisting of words from the MacArthur-Bates Communicative Development Inventories (CDIs)~\citep{fenson2007macarthur}. Following~\citet{portelance2020predicting}, we excluded any items comprising multiple words (such as ``choo-choo'') from our dataset, as the tokenizer would treat these as distinct items.
\end{itemize}
\vspace*{-8pt}
We select these two vocabulary sets as \texttt{CMN} offers broader coverage while \texttt{CDI} is more frequency-neutral and is used to assess children's early vocabulary development.
Following \citet{chang2022word}, we remove words with less than 100 occurrences in the evaluation set of each corpus to ensure reliable results, and keep at most 512 samples for each word.
This leads to 309 common words in \texttt{CMN}, as well as 345 words in \texttt{CDI} for BookCorpus and 243 words in \texttt{CDI} for BabyLM Corpus.

\vspace*{-5pt}
\paragraph{Surprisal and learning curves.}
In line with previous work, we use the mean surprisal (log-perplexity) of a word to quantify the quality of the model's predictions for this word.
For each occurrence of a word $w$, the surprisal is given by $-\log_2 P(w)$, and we average all occurrences.
To visualize word acquisition from high surprisal to lower surprisal, we evaluate the student model throughout training and plot the surprisal values over logarithm training steps, leading to a learning curve for individual words and the overall vocabulary.
We observe a similar pattern reported by \citet{chang2022word} that learning curves tend to level off at a local plateau, which aligns with the unigram surprisal.
This phenomenon renders the single-sigmoid model unreliable for capturing the complexity, and we adopt a double-sigmoid function to fit the learning curve.
We justify this choice and discuss more on the patterns in Appendix~\ref{sec:more-trial-examples}.

\input{sec/input/aoa-main}

\vspace*{-8pt}
\paragraph{Neural age of acquisition (nAoA).}
To evaluate the speed at which the student model acquires a word, we employ the neural age of acquisition (nAoA). 
Prior research~\citep{chang2022word,chang2023characterizing} has used a surprisal cutoff of 50\% between the minimum and maximum surprisal levels.
This is akin to the method used to determine children's age of acquisition, where the cutoff is set at the point when 50\% of children are observed to produce a word~\citep{braginsky2016from}.
To further enhance the robustness of this metric, we average nAoA over different surprisal thresholds from 0.5 to 0.95 with a step of 0.05, denoted as \texttt{nAoA@[0.50:0.95]}.
It's important to note that nAoA serves as a metric to assess the speed at which a model acquires a word, rather than the quality of word learning. 
A model might learn a word quickly, indicated by a low nAoA, yet not master it effectively, indicated by a high surprisal. 
Combining both metrics offers a thorough evaluation of word learning.

\input{sec/input/small-model}
\input{sec/input/mask}

\vspace*{-5pt}
\paragraph{Effective vocabulary size.}
Finally, we assess the effective vocabulary size relative to a test set of vocabulary. 
A word is deemed acquired at step $n$ if $\texttt{nAoA@0.50}\leq n$.
This approach yields a monotonically increasing curve that illustrates the growth of the effective vocabulary over time.

\vspace*{-5pt}
\section{Main Results and Findings}
\label{sec:main-results}

\vspace*{-5pt}
\subsection{Corrective feedback accelerates neural word acquisition}
\label{sec:results}

To evaluate \texttt{CMN} and \texttt{CDI} words on two corpora, we aggregate 5 random seeds and present the learning curves in Figure~\ref{fig:surprisal-main}, the neural age of acquisition in Figure~\ref{fig:aoa}, and the effective vocabulary size in Figure~\ref{fig:vocab}.
Figure~\ref{fig:surprisal-main} reveals that the \texttt{TnD} learning framework significantly accelerates word acquisition in training, outperforming other baselines. 
This acceleration is attributed to the critical roles of both trials and demonstrations in the learning process.
With only teacher demonstrations, the student model acquires words faster than with the plain \texttt{CLM} baseline alone, though not as rapidly as when active trials are incorporated in the \texttt{TnD} framework. 
Conversely, without the teacher's demonstrations, the student's trials in the wild do not yield a marked improvement, resulting in performance comparable to the \texttt{CLM} baseline.
We refer to Figure~\ref{fig:scatter-babylm} and~\ref{fig:scatter-bkps} in the Appendix for the ridgeline and scatter plot of words and their nAoA.

Figure~\ref{fig:aoa} and Table~\ref{tab:aoa} present nAoA at different surprisal thresholds from 0.50 to 0.95 with a step of 0.05.
We find that the \texttt{TnD} learning framework is particularly beneficial during the earlier stages of word acquisition, as it significantly outperforms the \texttt{CLM} baseline on \texttt{nAoA@0.50}, but is eventually on-par with the \texttt{CLM} baseline on \texttt{nAoA@0.95} towards the end of training.
As a result, it can be seen from Figure~\ref{fig:vocab} that students under the \texttt{TnD} framework quickly picked up a large volume of effective vocabulary, but eventually their vocabulary capacities have converged to the \texttt{CLM} baseline as expected. 
We further evaluate the final model on downstream natural language understanding (NLU) tasks.
The TnD model performs on par with the CLM model (See Appendix~\ref{sec:nlu}).
Overall, our findings show that corrective feedback accelerates the student model's neural word acquisition process, yet the student eventually converges to the teacher model's performance.
This contradicts the previous observations by \citet{nikolaus2021modeling} that combining production and perception-based language learning from the start will deteriorate performance.
We hypothesize that this discrepancy may result from using the BLEU score as a proxy for rewards from communicative feedback in their work, while our setup incorporates a more explicit form of corrective feedback.

\input{sec/input/aoa-masked}

\vspace*{-5pt}
\subsection{Corrective feedback helps knowledge distillation for smaller student models}

The current age-based reward design assumes that the student and teacher models are the same size. 
This section investigates whether such a reward function could distill linguistic knowledge to smaller student models.
The original student GPT-2 model has a dimension of $d=768$ (12 attention heads each with a dimension of 64). 
We now keep all experimental setups untouched but smaller student models with dimensions of 588 ($12\times49$), 360 ($10\times36$), and 250 ($10\times25$) respectively.
Figure~\ref{fig:smaller-models} shows that such efficient language learning can still be observed, even when the setting is translated to smaller models.
Each TnD model outperforms the CLM baseline of the same size and even surpasses CLM baselines of large capacity in early steps.

\vspace*{-5pt}
\subsection{Teacher's word preferences in demonstrations affect student}
\label{sec:teacher}

To explore how the teacher model's word selection impacts students' language development, we repeat the experiments where a chosen set of 40 words for each test vocabulary is excluded from teacher demonstrations. 
To ensure fluent generation, we maintain the presence of essential functional words so these words don't appear in the 40 chosen words.
During the language generation process by the teacher model, if a word from this set is to be decoded, we select the next best alternative, ensuring these words were never presented in teacher demonstrations.
We show the learning curves for these excluded words in Figure~\ref{fig:mask} and present the nAoA in Table~\ref{tab:aoa-masked} (in Appendix).
Our findings indicate that the teacher model's word choices significantly influence the efficiency of word acquisition by the student model. 
The absence of words from teacher demonstrations leads to slower learning speed for student models, as evidenced by a higher \texttt{nAoA}, although the student models are ultimately able to learn these words from the corpus and their trials.

\input{sec/input/trial}

\vspace*{-5pt}
\subsection{Practice makes perfect in trials}

Finally, we conduct experiments to underscore the significance of the student's active trials in the process of word learning.
A student model can learn a word from 3 sources: its own trials, teacher demonstrations, and exposure to the corpus. 
To determine which source contributes most significantly to learning, we begin by plotting the per-word learning curves alongside the cumulative frequency of word encounters in trials, demonstrations, and the corpus.
We observe, interestingly, that the learning curves for certain words exhibit a pronounced correlation with the frequency of these words in trials, as exampled in Figure~\ref{fig:trial} and~\ref{fig:more-trial}.

We speculate that this pattern may be associated with the part of speech (POS) of the word.
Following~\citet{portelance2020predicting}, we delve into this phenomenon by focusing on specific subsets of words, including nouns, predicates, and functional words, to explore the relationship further.
To evaluate the impact of each source of word acquisition, we consider the cumulative word frequency as a predictor of the word surprisal, and carry out linear regressions complemented by likelihood ratio tests to determine the beta weights for each predictor. 
Upon analysis, we identify significant collinearity between word frequency in teacher demonstrations and the corpus, as indicated by a high variance inflation factor (VIF), while the word frequency in student trials is less intertwined, exhibiting a moderate VIF below 10.
We thus calculate the beta weights separately for pairs (trial, demo) and (trial, corpus), then compute the average beta weights for trials.
Together with the Pearson correlation, we summarize the results in Table~\ref{tab:predictor}.
Negative beta weights signify a negative correlation, with a larger absolute value denoting a stronger association and contribution.
Our analysis reveals that the cumulative frequency of words encountered in trials plays a significant role in the acquisition of functional words and predicates. However, this significant contribution does not extend to nouns, indicating a potential impact of active trials on different POS within the learning process.
This finding is linguistically intuitive, as function words and predicates are words that require other dependent words to fully express their meaning~\citep{gleitman1990structural} and thus require more practice. 
We posit that grounding language in the non-linguistic world is essential for acquiring the meanings of words, particularly for concrete noun~\citep{ma2023world}.

\input{sec/input/curve-predictor}

%% file: sec/input/surprisal-main.tex
\begin{figure*}[!h]
    \centering
    \begin{subfigure}[t]{.23\textwidth}
        \centering
        \includegraphics[width=1.05\columnwidth]{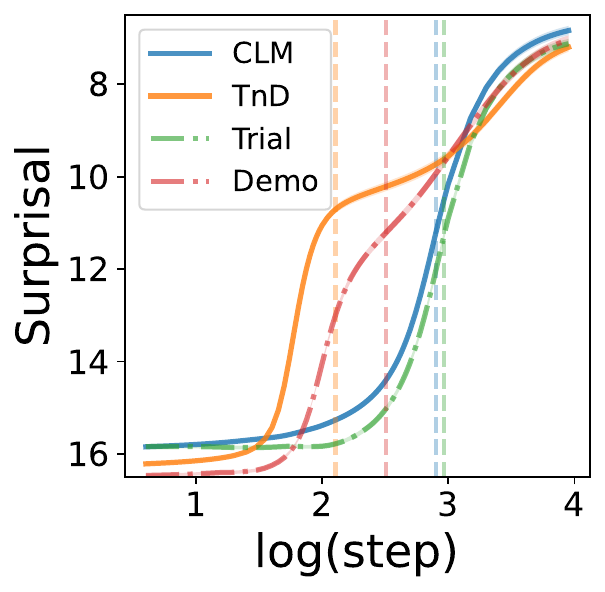}
        \vspace*{-20pt}
        \caption{Learning curve of \texttt{CMN} words on BookCorpus.}
        \label{fig:sps-bkps-cmn}
    \end{subfigure}
    ~
    \begin{subfigure}[t]{.23\textwidth}
        \centering
        \includegraphics[width=1.05\columnwidth]{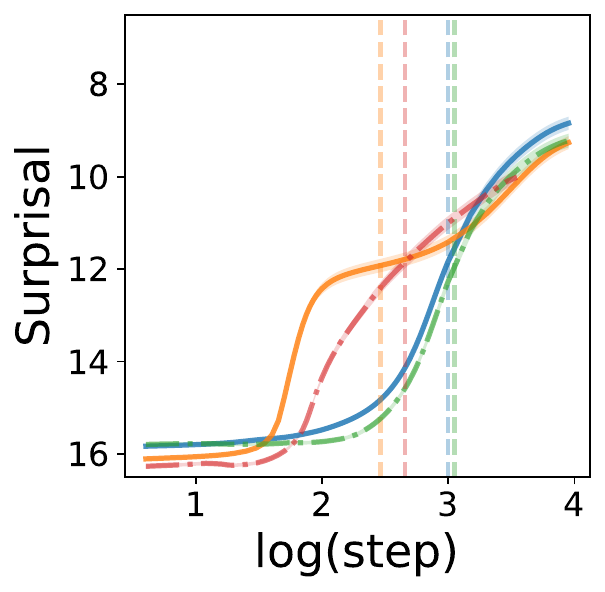}
        \vspace*{-20pt}
        \caption{Learning curve of \texttt{CDI} words on BookCorpus.}
        \label{fig:sps-bkps-cdi}
    \end{subfigure}
    ~
    \begin{subfigure}[t]{.23\textwidth}
        \centering
        \includegraphics[width=1.05\columnwidth]{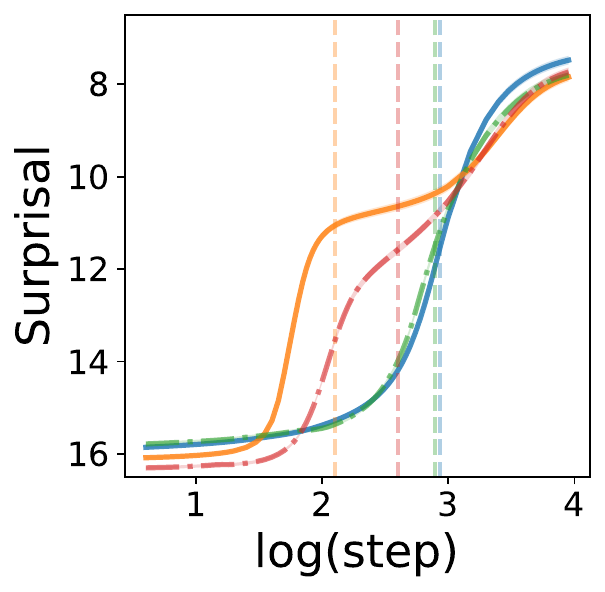}
        \vspace*{-20pt}
        \caption{Learning curve of \texttt{CMN} words on BabyLM.}
        \label{fig:sps-bblm-cmn}
    \end{subfigure}
    ~
    \begin{subfigure}[t]{.23\textwidth}
        \centering
        \includegraphics[width=1.05\columnwidth]{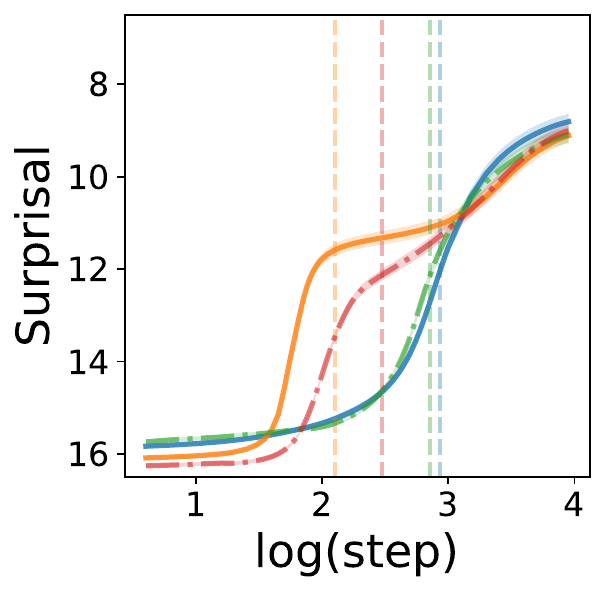}
        \vspace*{-20pt}
        \caption{Learning curve of \texttt{CDI} words on BabyLM.}
        \label{fig:sps-bblm-cdi}
    \end{subfigure}
    \vspace*{-5pt}
    \caption{On 2 training corpora and 2 test vocabulary, we aggregate 5 random seeds and present the fitted learning curves of mean surprisal over $\log_{10}$ training steps, with \texttt{nAoA@0.5} of each curve indicated by a vertical dashed line. \vspace*{-10pt}}
    \label{fig:surprisal-main}
\end{figure*}

%% file: sec/input/aoa-figure.tex
\begin{figure*}[!t]
    \centering
    \begin{subfigure}[t]{.23\textwidth}
        \centering
        \includegraphics[width=1.05\linewidth]{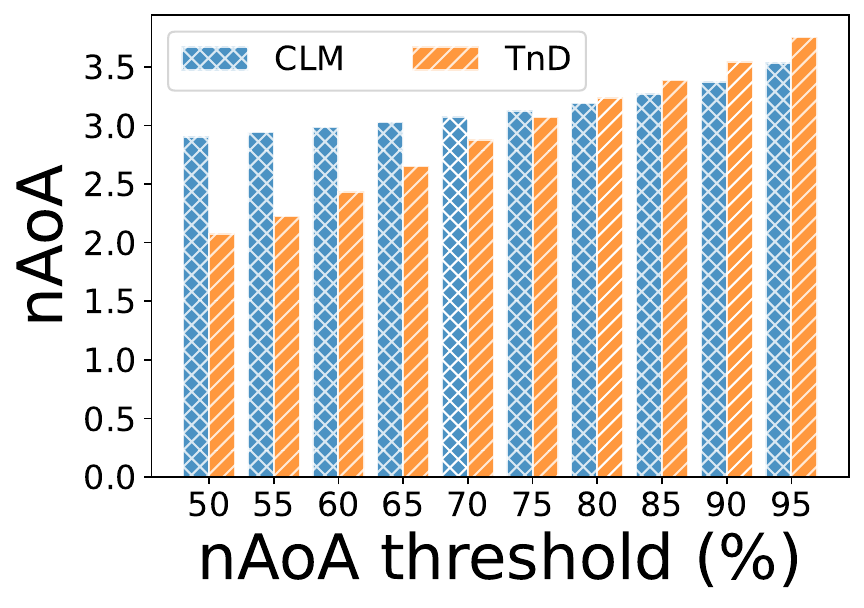}
        \vspace*{-18pt}
        \caption{nAoA of \texttt{CMN} words on BookCorpus.}
        \label{fig:aoa-bblm-cmn}
    \end{subfigure}
    ~
    \begin{subfigure}[t]{.23\textwidth}
        \centering
        \includegraphics[width=1.05\linewidth]{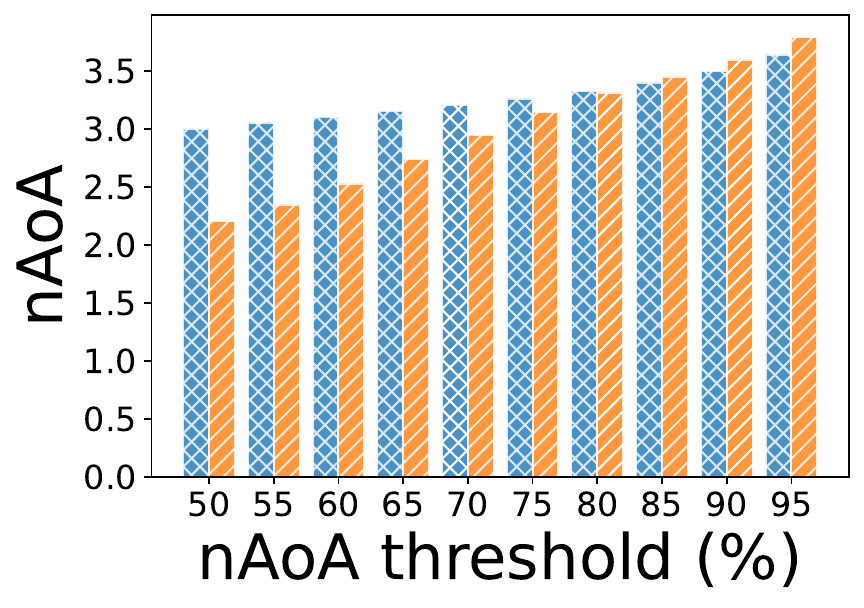}
        \vspace*{-18pt}
        \caption{nAoA of \texttt{CDI} words on BookCorpus.}
        \label{fig:aoa-bblm-cmn}
    \end{subfigure}
    ~
    \begin{subfigure}[t]{.23\textwidth}
        \centering
        \includegraphics[width=1.05\linewidth]{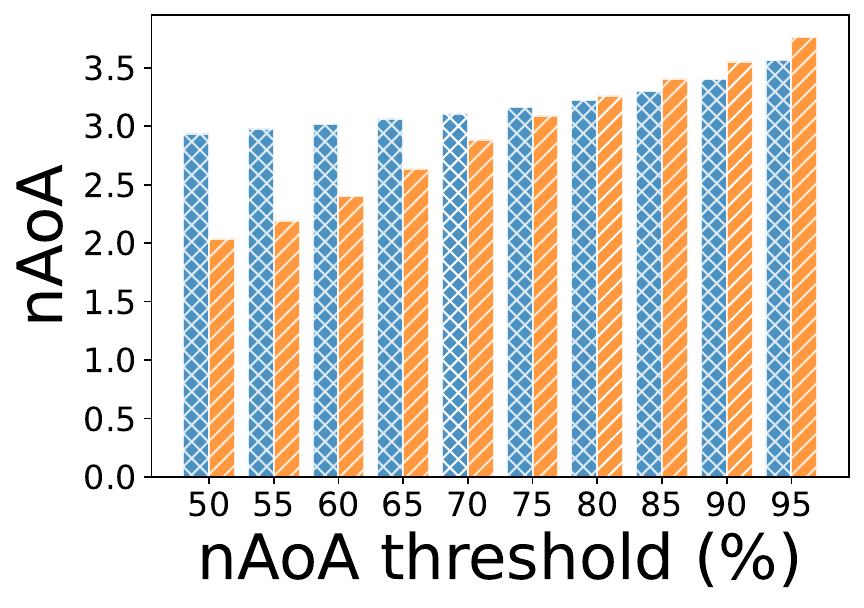}
        \vspace*{-18pt}
        \caption{nAoA of \texttt{CMN} words on BabyLM.}
        \label{fig:aoa-bblm-cmn}
    \end{subfigure}
    ~
    \begin{subfigure}[t]{.23\textwidth}
        \centering
        \includegraphics[width=1.05\linewidth]{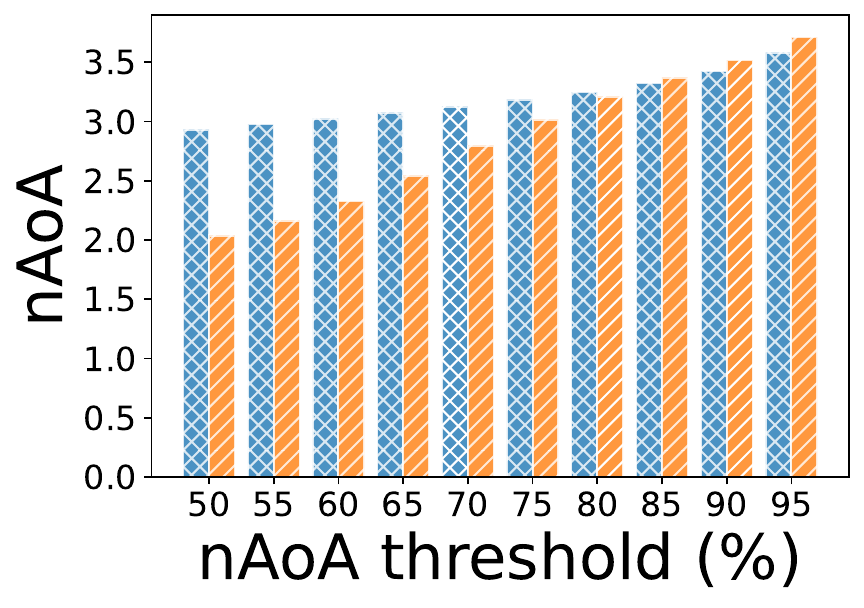}
        \vspace*{-18pt}
        \caption{nAoA of \texttt{CDI} words on BabyLM.}
        \label{fig:aoa-bblm-cmn}
    \end{subfigure}
    \vspace*{-9pt}
    \caption{On 2 training corpora and 2 test vocabulary, we aggregate 5 random seeds and present the neural age of acquisition (nAoA) at different surprisal thresholds from 0.5 to 0.95 with a step of 0.05. \vspace*{-26pt}}
    \label{fig:aoa}
\end{figure*}

%% file: sec/input/vocab.tex
\begin{figure*}[!t]
    \vspace{15pt}
    \centering
    \begin{subfigure}[t]{.23\textwidth}
        \centering
        \includegraphics[width=1.05\linewidth]{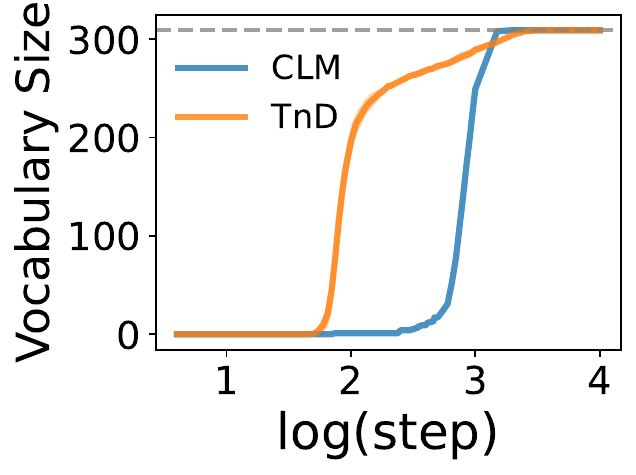}
        \vspace*{-20pt}
        \caption{Effective vocabulary size in \texttt{CMN} words on BookCorpus.}
        \label{fig:vocab-bblm-cmn}
    \end{subfigure}
    ~
    \begin{subfigure}[t]{.23\textwidth}
        \centering
        \includegraphics[width=1.05\linewidth]{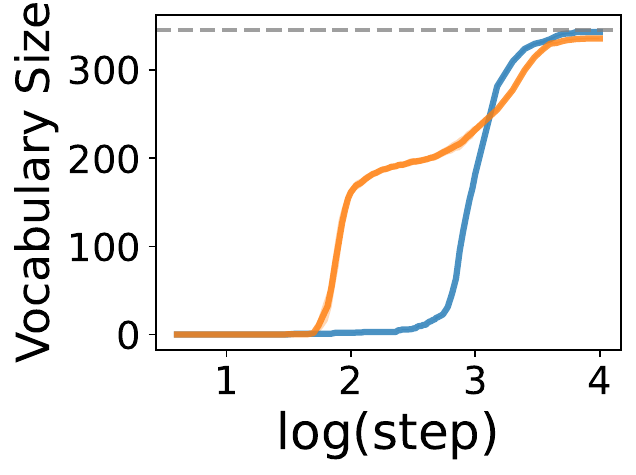}
        \vspace*{-20pt}
        \caption{Effective vocabulary size in \texttt{CDI} words on BookCorpus.}
        \label{fig:vocab-bblm-cmn}
    \end{subfigure}
    ~
    \begin{subfigure}[t]{.23\textwidth}
        \centering
        \includegraphics[width=1.05\linewidth]{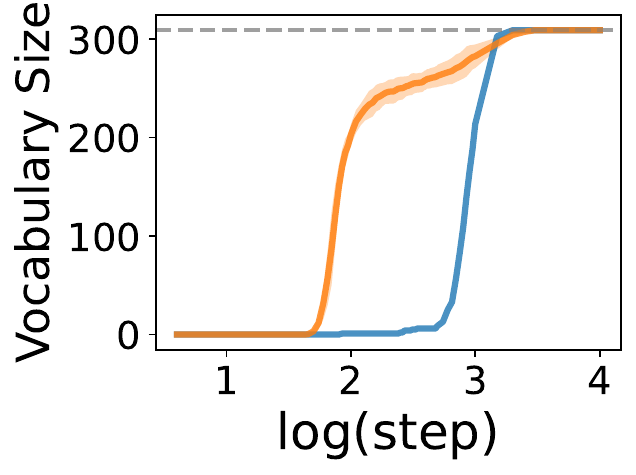}
        \vspace*{-20pt}
        \caption{Effective vocabulary size in \texttt{CMN} words on BabyLM.}
        \label{fig:vocab-bblm-cmn}
    \end{subfigure}
    ~
    \begin{subfigure}[t]{.23\textwidth}
        \centering
        \includegraphics[width=1.05\linewidth]{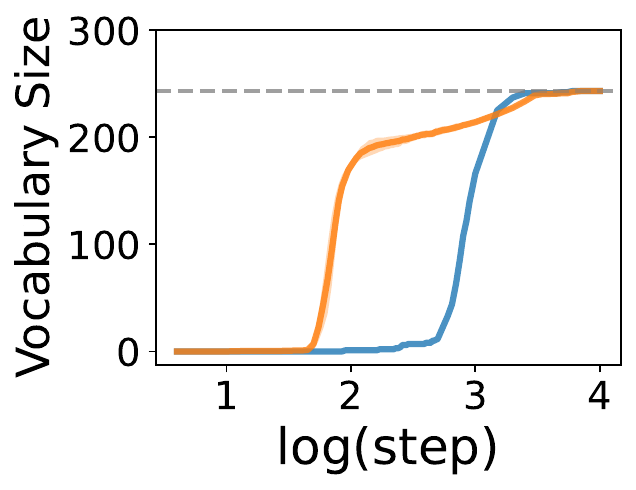}
        \vspace*{-20pt}
        \caption{Effective vocabulary size in \texttt{CDI} words on BabyLM.}
        \label{fig:vocab-bblm-cmn}
    \end{subfigure}
    \vspace*{-8pt}
    \caption{On 2 training corpora and 2 test vocabulary, we aggregate 5 random seeds and evaluate the effective vocabulary size over $\log_{10}$ training steps. The dashed lines mark the tested vocabulary size. \vspace*{-18pt}}
    \label{fig:vocab}
\end{figure*}

%% file: sec/input/aoa-main.tex
\begin{table}[!t]
\vspace{3pt}
\centering
    \scalebox{0.80}{
    \begingroup
    \setlength{\tabcolsep}{1.9pt}
    \renewcommand{\arraystretch}{0.8}
    \hspace*{-10pt}
    \begin{tabular}{cccccc}
    \toprule
    \multirow{2}{*}{Method} & \multirow{2}{*}{\texttt{nAoA}$\downarrow$} & \multicolumn{2}{c}{BabyLM Corpus} & \multicolumn{2}{c}{BookCorpus} \\
    \cmidrule(r){3-4} \cmidrule(r){5-6}
      &   & \texttt{CMN}          & \texttt{CDI}         & \texttt{CMN}            & \texttt{CDI}           \\
    \cmidrule(r){1-1} \cmidrule(r){2-2} \cmidrule(r){3-4} \cmidrule(r){5-6}
    \multirow{2}{*}{\texttt{CLM}} &  \texttt{@.5}                & $2.94_{\pm0.01}$         & $2.93_{\pm0.01}$        & $2.90_{\pm0.01}$           & $3.00_{\pm0.01}$          \\
                         &   \texttt{@[.5:.95]}        & $3.18_{\pm0.01}$         & $3.19_{\pm0.01}$        & $3.14_{\pm0.01}$           & $3.26_{\pm0.01}$         \\
    \cmidrule(r){1-1} \cmidrule(r){2-2} \cmidrule(r){3-4} \cmidrule(r){5-6}
    \multirow{2}{*}{\texttt{Trial}} &  \texttt{@.5}                & $2.90_{\pm0.01}$         & $2.86_{\pm0.01}$        & $2.97_{\pm0.01}$           & $3.05_{\pm0.01}$          \\
                     &   \texttt{@[.5:.95]}        & $3.20_{\pm0.01}$         & $3.17_{\pm0.01}$        & $3.21_{\pm0.01}$           & $3.30_{\pm0.01}$         \\
    \cmidrule(r){1-1} \cmidrule(r){2-2} \cmidrule(r){3-4} \cmidrule(r){5-6}
    \multirow{2}{*}{\texttt{Demo}} &  \texttt{@.5}                & $2.60_{\pm0.02}$         & $2.47_{\pm0.03}$        & $2.51_{\pm0.02}$           & $2.66_{\pm0.02}$          \\
                     &   \texttt{@[.5:.95]}        & $3.15_{\pm0.01}$         & $3.08_{\pm0.02}$        & $3.05_{\pm0.01}$           & $\textbf{2.97}_{\pm0.02}$         \\
    \cmidrule(r){1-1} \cmidrule(r){2-2} \cmidrule(r){3-4} \cmidrule(r){5-6}
    \multirow{2}{*}{\texttt{TnD}} &  \texttt{@.5}                & $\textbf{2.10}_{\pm0.02}$         & $\textbf{2.10}_{\pm0.03}$        & $\textbf{2.11}_{\pm0.02}$           & $\textbf{2.46}_{\pm0.03}$          \\
                     &   \texttt{@[.5:.95]}        & $\textbf{2.95}_{\pm0.02}$         & $\textbf{2.87}_{\pm0.03}$        & $\textbf{2.90}_{\pm0.02}$           & $3.11_{\pm0.02}$         \\
    \bottomrule
    \end{tabular}
    \endgroup}
\vspace*{-5pt}
\caption{For each baseline and setup, we report the neural age of acquisition (\texttt{nAoA}) with standard errors at 0.50 cutoff and averaged over surprisal thresholds from 0.50 to 0.95 with a step of 0.05. 
\vspace*{-20pt}
}
\label{tab:aoa}
\end{table}

%% file: sec/input/small-model.tex
\begin{figure*}[!t]
    \vspace{15pt}
    \centering
    \begin{subfigure}[t]{1\textwidth}
        \centering
        \includegraphics[width=1.0\linewidth]{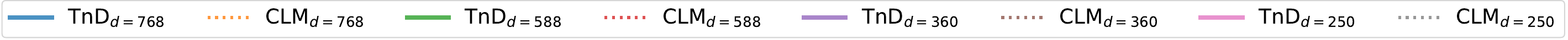}
        \vspace*{-20pt}
    \end{subfigure}
    ~
    \begin{subfigure}[t]{.23\textwidth}
        \centering
        \includegraphics[width=1.05\linewidth]{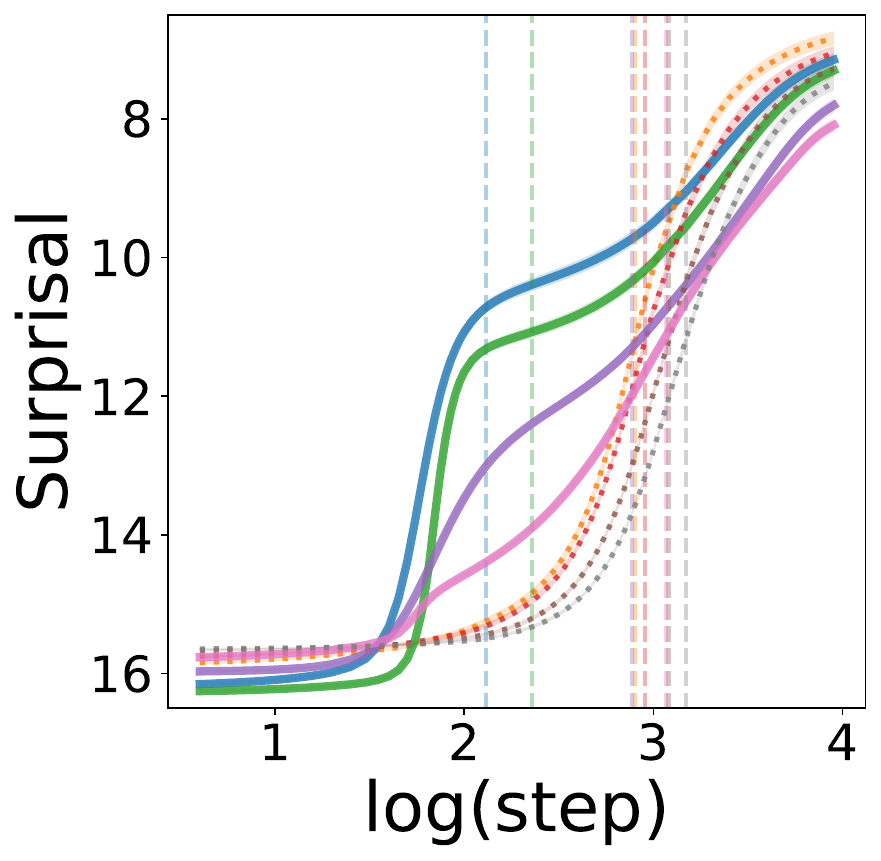}
        \vspace*{-20pt}
        \caption{Learning curve sof smaller models of \texttt{CMN} words on BookCorpus.}
        \label{fig:smaller-bblm-cmn}
    \end{subfigure}
    ~
    \begin{subfigure}[t]{.23\textwidth}
        \centering
        \includegraphics[width=1.05\linewidth]{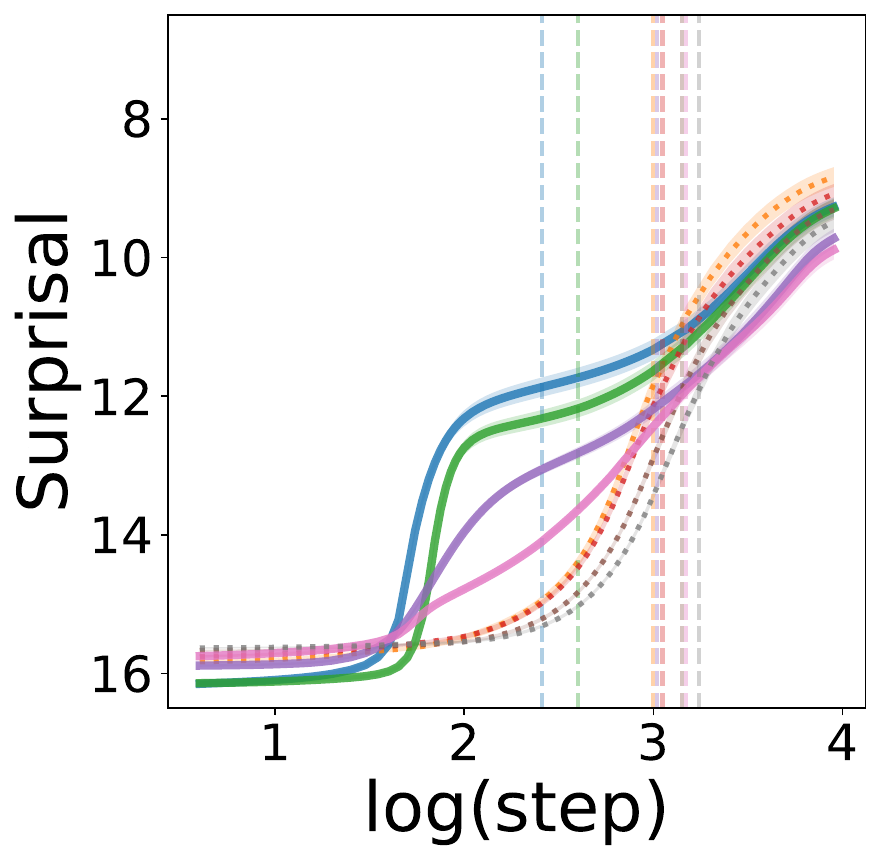}
        \vspace*{-20pt}
        \caption{Learning curve sof smaller models of \texttt{CDI} words on BookCorpus.}
        \label{fig:smaller-bblm-cmn}
    \end{subfigure}
    ~
    \begin{subfigure}[t]{.23\textwidth}
        \centering
        \includegraphics[width=1.05\linewidth]{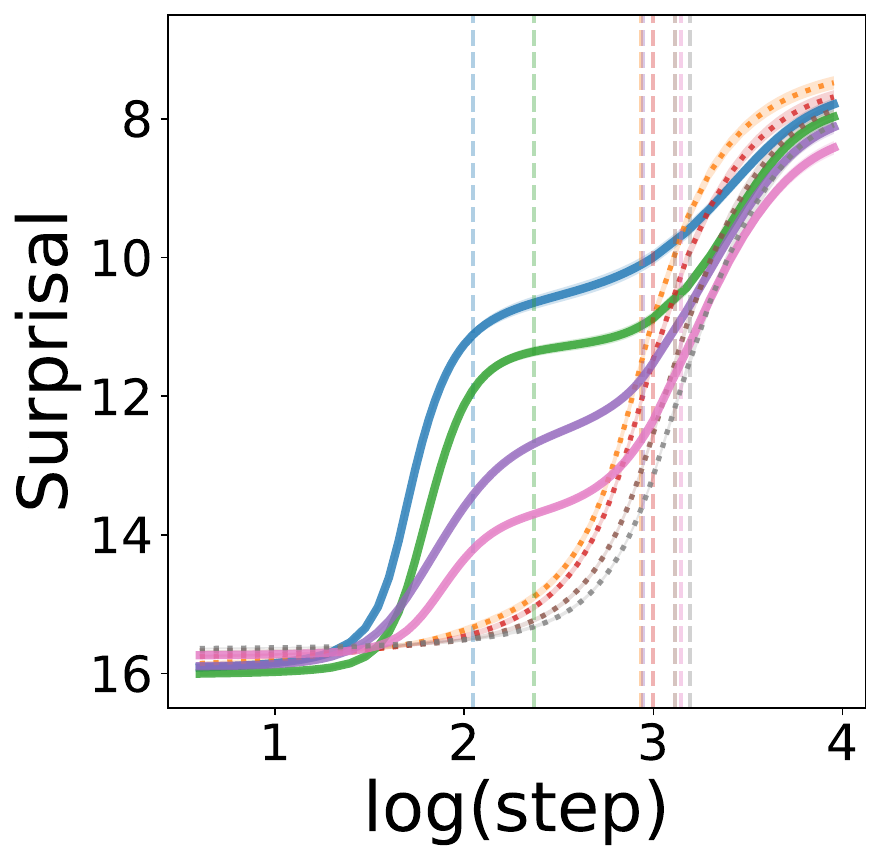}
        \vspace*{-20pt}
        \caption{Learning curve sof smaller models of \texttt{CMN} words on BabyLM.}
        \label{fig:smaller-bblm-cmn}
    \end{subfigure}
    ~
    \begin{subfigure}[t]{.23\textwidth}
        \centering
        \includegraphics[width=1.05\linewidth]{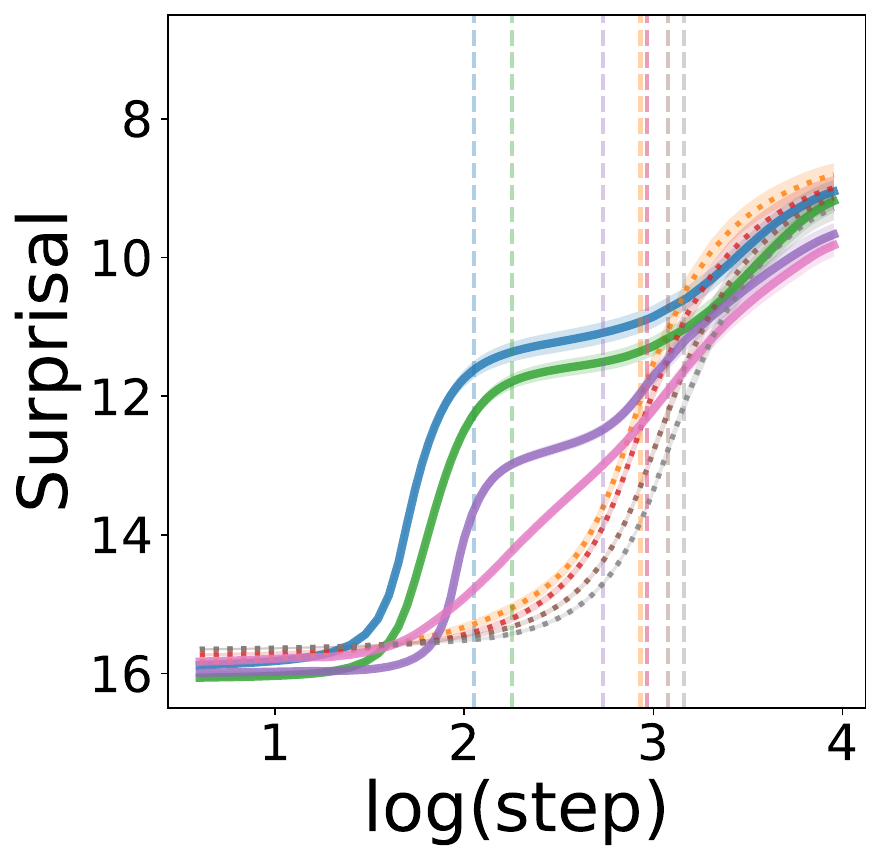}
        \vspace*{-20pt}
        \caption{Learning curve sof smaller models of \texttt{CDI} words on BabyLM.}
        \label{fig:smaller-bblm-cmn}
    \end{subfigure}
    \vspace*{-9pt}
    \caption{We further present the fitted learning curves on smaller student models ($\texttt{TnD}_{d=588/360/250}$) of mean surprisal over $\log_{10}$ training steps, with $\texttt{nAoA@0.5}$ of each curve indicated by a vertical dashed line. \vspace*{-10pt}}
    \label{fig:smaller-models}
\end{figure*}

%% file: sec/input/mask.tex
\begin{figure*}[!t]
    \centering
    \begin{subfigure}[t]{.23\textwidth}
        \centering
        \includegraphics[width=1.05\linewidth]{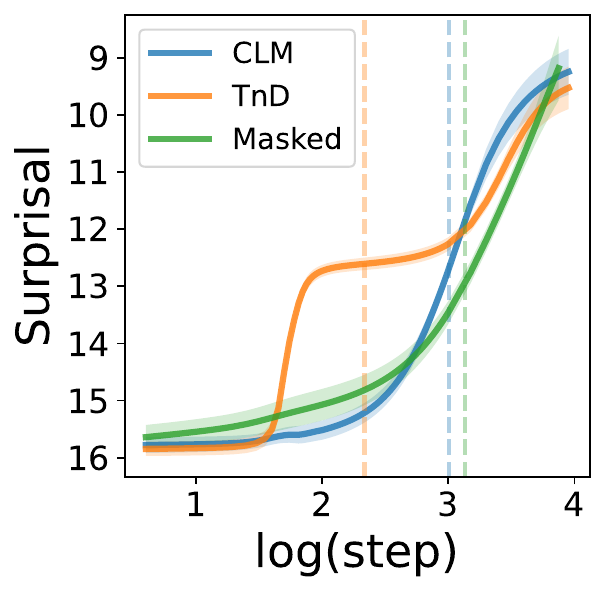}
        \vspace*{-20pt}
        \caption{Influence of teacher's word preferences in \texttt{CMN} words on BabyLM.}
        \label{fig:mask-bblm-cmn}
    \end{subfigure}
    ~
    \begin{subfigure}[t]{.23\textwidth}
        \centering
        \includegraphics[width=1.05\linewidth]{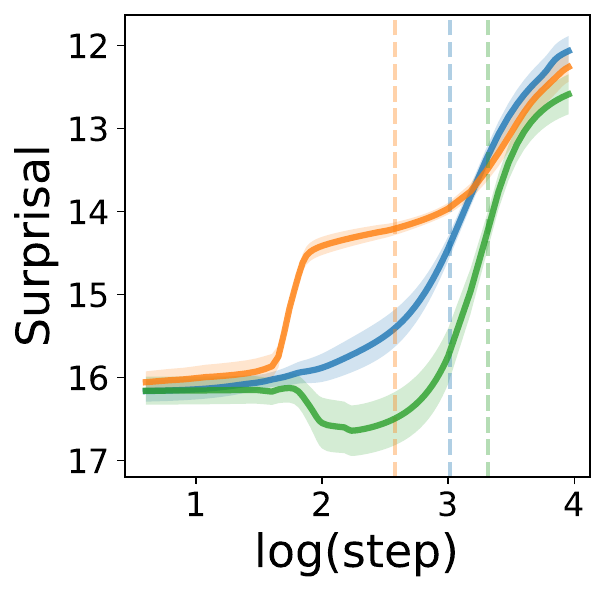}
        \vspace*{-20pt}
        \caption{Influence of teacher's word preferences in \texttt{CDI} words on BabyLM.}
        \label{fig:mask-bblm-cmn}
    \end{subfigure}
    ~
    \begin{subfigure}[t]{.23\textwidth}
        \centering
        \includegraphics[width=1.05\linewidth]{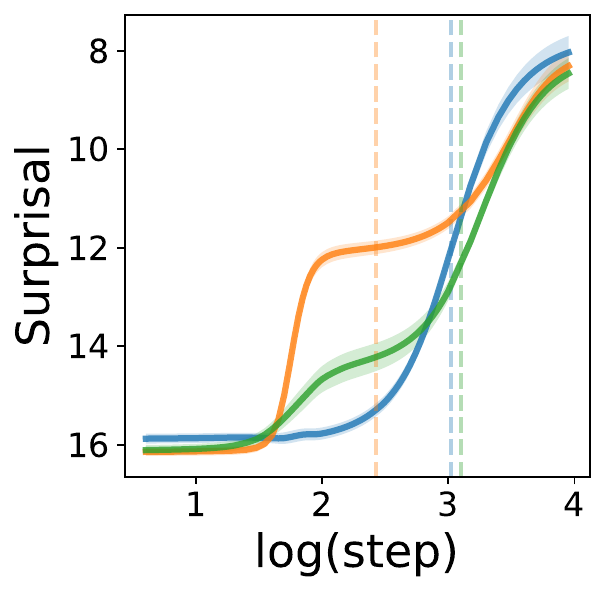}
        \vspace*{-20pt}
        \caption{Influence of teacher's word preferences in \texttt{CMN} words on BookCorpus.}
        \label{fig:mask-bblm-cmn}
    \end{subfigure}
    ~
    \begin{subfigure}[t]{.23\textwidth}
        \centering
        \includegraphics[width=1.05\linewidth]{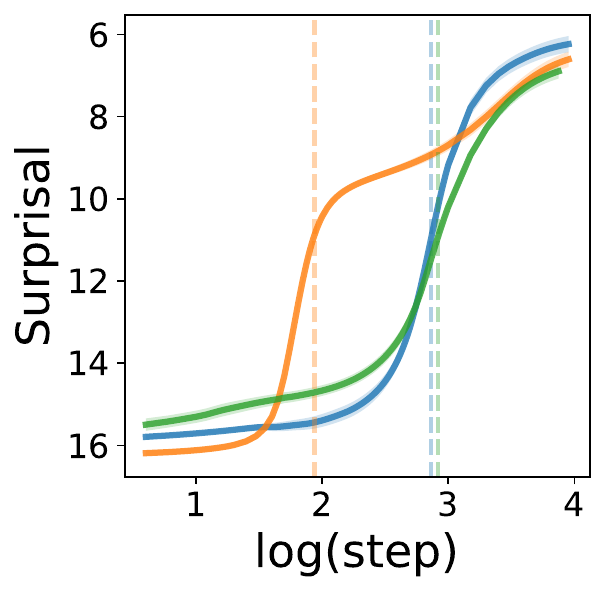}
        \vspace*{-20pt}
        \caption{Influence of teacher's word preferences in \texttt{CDI} words on BookCorpus.}
        \label{fig:mask-bblm-cmn}
    \end{subfigure}
    \vspace*{-5pt}
    \caption{For the 40 words to be ``masked out'' from teacher demonstrations, we repeat the experiment with 5 random seeds and plot the learning curves of these words with those in \texttt{CLM} and \texttt{TnD} baselines.\vspace*{-15pt}}
    \label{fig:mask}
\end{figure*}

%% file: sec/input/aoa-masked.tex
\begin{table}[!t]
\vspace{5pt}
\centering
    \scalebox{0.775}{
    \begingroup
    \setlength{\tabcolsep}{2pt}
    \renewcommand{\arraystretch}{0.8}
    \hspace*{-10pt}
    \begin{tabular}{cccccc}
    \toprule
    \multirow{2}{*}{Method} & \multirow{2}{*}{\texttt{nAoA}$\downarrow$} & \multicolumn{2}{c}{BabyLM Corpus} & \multicolumn{2}{c}{BookCorpus} \\
    \cmidrule(r){3-4} \cmidrule(r){5-6}
      &   & \texttt{CMN}          & \texttt{CDI}         & \texttt{CMN}            & \texttt{CDI}           \\
    \cmidrule(r){1-1} \cmidrule(r){2-2} \cmidrule(r){3-4} \cmidrule(r){5-6}
    \multirow{2}{*}{\texttt{CLM}} &  \texttt{@.5}                & $3.01_{\pm0.03}$         & $3.02_{\pm0.07}$        & $3.02_{\pm0.02}$           & $2.86_{\pm0.01}$          \\
                         &   \texttt{@[.5:.95]}        & $3.29_{\pm0.02}$         & $3.37_{\pm0.03}$        & $3.28_{\pm0.02}$           & $3.09_{\pm0.01}$         \\
    \cmidrule(r){1-1} \cmidrule(r){2-2} \cmidrule(r){3-4} \cmidrule(r){5-6}
    \multirow{2}{*}{\texttt{TnD}} &  \texttt{@.5}                & $\textbf{2.34}_{\pm0.15}$         & $\textbf{2.57}_{\pm0.14}$        & $\textbf{2.43}_{\pm0.12}$           & $\textbf{1.94}_{\pm0.02}$          \\
                     &   \texttt{@[.5:.95]}        & $\textbf{3.18}_{\pm0.07}$         & $\textbf{3.22}_{\pm0.08}$        & $\textbf{3.22}_{\pm0.05}$           & $\textbf{2.76}_{\pm0.04}$         \\
    \cmidrule(r){1-1} \cmidrule(r){2-2} \cmidrule(r){3-4} \cmidrule(r){5-6}
    \texttt{TnD} &  \texttt{@.5}                & \textcolor{red}{$3.14_{\pm0.09}$}         & \textcolor{red}{$3.32_{\pm0.04}$}        & \textcolor{red}{$3.10_{\pm0.04}$}           & \textcolor{red}{$2.92_{\pm0.02}$}          \\
    (Masked) &   \texttt{@[.5:.95]}        & \textcolor{red}{$3.49_{\pm0.03}$}         & \textcolor{red}{$3.50_{\pm0.03}$}        & \textcolor{red}{$3.42_{\pm0.02}$}           & \textcolor{red}{$3.21_{\pm0.02}$}         \\
    \bottomrule
    \end{tabular}
    \endgroup}
\vspace*{-5pt}
\caption{For the 40 words to be ``masked out'' from teacher demonstrations, we repeat the experiment with 5 random seeds and compared their \texttt{nAoA} with standard errors to those observed in \texttt{CLM} and \texttt{TnD} baselines. \vspace*{-15pt}}
\label{tab:aoa-masked}
\end{table}

%% file: sec/input/trial.tex
\begin{figure}[!t]
    \centering
    \begin{subfigure}[t]{.23\textwidth}
        \centering
        \includegraphics[width=1.05\linewidth]{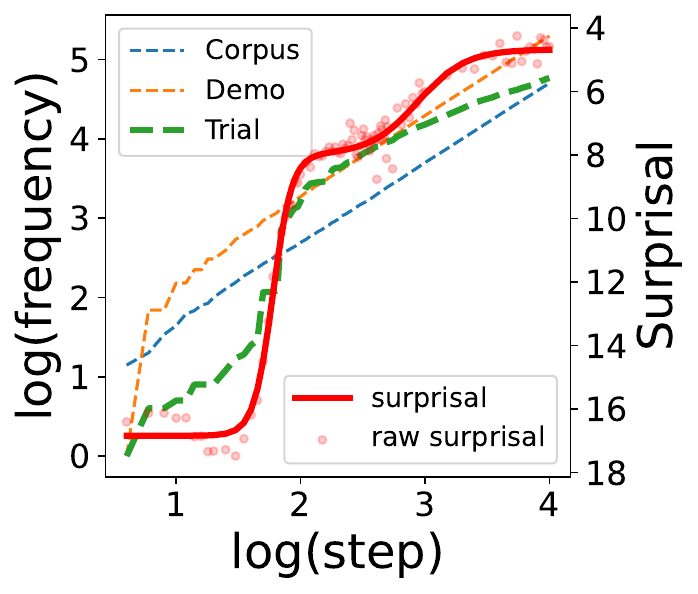}
        \vspace*{-20pt}
        \caption{The word ``have.''}
    \end{subfigure}
    ~
    \begin{subfigure}[t]{.23\textwidth}
        \centering
        \includegraphics[width=1.05\linewidth]{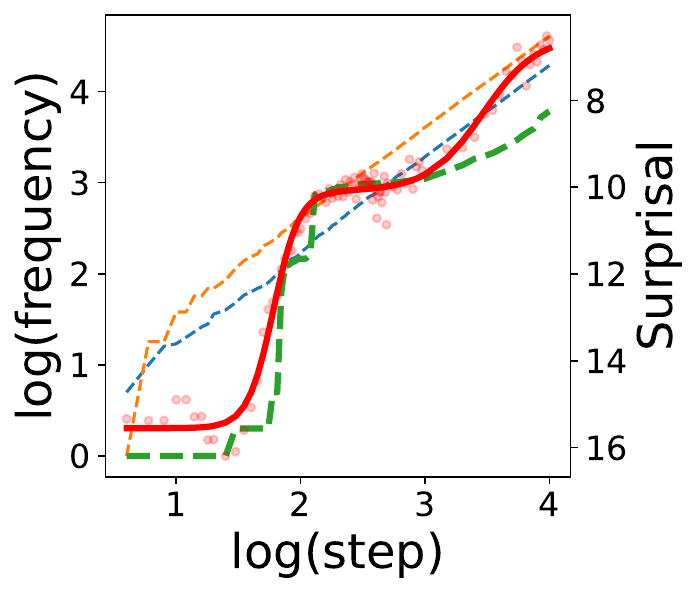}
        \vspace*{-20pt}
        \caption{The word ``now.''}
    \end{subfigure}
    \vspace*{-10pt}
    \caption{Examples of per-word learning curves and cumulative word frequency in BookCorpus. The dashed lines mark the log frequency of words (left $y$-axis) from each source. The solid line and dots mark the word surprisal (right $y$-axis).\vspace*{-15pt}}
    \label{fig:trial}
\end{figure}

%% file: sec/input/curve-predictor.tex
\begin{table}[!t]
    \centering
        \scalebox{0.8}{
        \begingroup
        \setlength{\tabcolsep}{2pt}
        \renewcommand{\arraystretch}{0.8}
        \hspace*{-10pt}
        \begin{tabular}{cccccccccc}
        \toprule
        \multirow{3}{*}{POS} & \multirow{3}{*}{Freq.} & \multicolumn{4}{c}{BabyLM Corpus}                            & \multicolumn{4}{c}{BookCorpus}                        \\
        \cmidrule(r){3-6} \cmidrule(r){7-10}
                             &                            & \multicolumn{2}{c}{\texttt{CMN}}   & \multicolumn{2}{c}{\texttt{CDI}}   & \multicolumn{2}{c}{\texttt{CMN}}   & \multicolumn{2}{c}{\texttt{CDI}}   \\
        \cmidrule(r){3-4} \cmidrule(r){5-6} \cmidrule(r){7-8} \cmidrule(r){9-10}
                             &                            & $\beta$ & $r$ & $\beta$ & $r$ & $\beta$ & $r$ & $\beta$ & $r$ \\
        \cmidrule(r){1-1} \cmidrule(r){2-2} \cmidrule(r){3-4} \cmidrule(r){5-6} \cmidrule(r){7-8} \cmidrule(r){9-10}
        \multirow{3}{*}{noun}       & trial  & -0.36 & -0.90 & -0.25 & -0.85 & -0.38 & -0.92 & -0.31 & -0.85 \\
                                    & demo   & \textbf{-0.67} & -0.93 & \textbf{-0.73} & -0.89 & \textbf{-0.56} & -0.93 & \textbf{-0.67} & -0.87 \\
                                    & corpus & -0.51 & -0.93 & -0.53 & -0.88 & \textbf{-0.56} & -0.93 & -0.60 & -0.87 \\
        \cmidrule(r){1-1} \cmidrule(r){2-2} \cmidrule(r){3-4} \cmidrule(r){5-6} \cmidrule(r){7-8} \cmidrule(r){9-10}
        \multirow{3}{*}{pred}  & trial  & \textbf{-0.70} & -0.90 & \textbf{-0.72} & -0.86 & \textbf{-0.49} & -0.92 & \textbf{-0.54} & -0.88 \\
                                    & demo   & -0.33 & -0.93 & -0.30 & -0.90 & \textbf{-0.49} & -0.92 & -0.45 & -0.90 \\
                                    & corpus & -0.22 & -0.93 & -0.19 & -0.90 & -0.45 & -0.93 & -0.40 & -0.87 \\
        \cmidrule(r){1-1} \cmidrule(r){2-2} \cmidrule(r){3-4} \cmidrule(r){5-6} \cmidrule(r){7-8} \cmidrule(r){9-10}
        \multirow{3}{*}{func} & trial  & \textbf{-0.67} & -0.93 & \textbf{-0.72} & -0.92 & \textbf{-0.67} & -0.94 & \textbf{-0.59} & -0.93 \\
                                    & demo   & -0.39 & -0.92 & -0.21 & -0.90 & -0.22 & -0.91 & -0.37 & -0.87 \\
                                    & corpus & -0.17 & -0.92 & -0.25 & -0.91 & -0.35 & -0.92 & -0.35 & -0.90  \\
        \bottomrule
        \end{tabular}
    \endgroup}
    \vspace{-5pt}
    \caption{For each POS category, we present the beta weights $\beta$ and Pearson correlation $r$ between their mean surprisal and cumulative word frequency over the course of training. These metrics are evaluated based on the student's trials, teacher's demonstrations, and the overall corpus frequency up to the current training step. \vspace{-15pt}}
    \label{tab:predictor}
\end{table}

%% file: sec/02-background.tex
\vspace*{-5pt}
\section{Related Work and Discussions}
\label{sec:related-word}

\vspace*{-5pt}
\subsection{Interaction in neural language learning}

Researchers have emphasized the role of interaction in computational models of language~\citep{bisk2020experience,tsuji2021scala}.
Preliminary efforts have been conducted under specific constraints, such as in domain-specific scenarios~\citep{qu2010context,weston2016dialog,bianchi2021language,stein2021shapelurn,madotto2021exploration} or considering particular types of dialogue acts~\citep{zhang2018interactive,yuan2019interactive}.
More recently, a series of studies have approached language acquisition through the lens of multimodal referential games~\citep{lazaridou2016towards,zhu2022language,liu2022computational}, emphasizing the importance of pragmatic inference and communicative feedback in speaker-listener interactions.
Early work investigated scenarios where the teacher models actively select training data to optimally assist a passive student learner~\citep{ter2022towards}.
\citet{nikolaus2021modeling} adopted setups where student models learn by producing language and receive feedback of communicative success using the BLEU score~\citep{papineni2002bleu}.
However, these models do not receive teachers' demonstrations or corrections.
Our work diverges from these studies as we focus on symmetric teacher-student interactions that require language production from the student and explicit corrective feedback from the teacher, which goes beyond the simple speaker-listener roles.
We also adopt massive corpora to study modern transformer-based generative language models, a general approach without domain restrictions.
Recent advancements in reinforcement learning from human feedback (RLHF)~\citep{ouyang2022training} mark a breakthrough in interactive language learning.
This work leverages the implementation infrastructure of RLHF, but diverges in significant ways.
Whereas RLHF aims to align a pre-trained language model with human preferences, our objective is to ``babysit'' a language model from scratch.
We refer to existing surveys~\citep{kaufmann2023survey,zheng2023secrets} and discuss RLHF further in Appendix~\ref{sec:rlhf-app}.

\vspace*{-5pt}
\subsection{Psychologically motivated analysis of language models}
\label{sec:psychlm-app}

Although language models and human language learners differ in their inductive biases and data sources~\citep{baroni2022proper,warstadt2022artificial}, several works have looked into the learnability, proficiency, and efficiency of neural language learning, for example, the relationships between word surprisal in language models to various psycholinguistic variables~\citep{portelance2020predicting,ma2023world}.
Recent efforts have shifted towards exploring the developmental trajectories of language models, rather than their end performance~\citep{sellam2021multiberts,blevins2022analyzing,biderman2023pythia,xia2023training}, sparkling further investigations into the developmental aspects of psycholinguistics using computational approaches~\citep{chang2022word,chang2023characterizing,evanson2023language}.
The scientific rationale behind this is that these models can serve as hypothesis generators or proofs of concept, verifying mechanisms that are practically effective for machines and potentially feasible for human learners~\citep{portelance2022neural,portelance2023roles}.
We echo that the benchmark outcomes of language models themselves are insufficient~\cite{baroni2022proper,portelance2022neural}, and researchers need to control the factors that may have contributed to the model's learning process, particularly through conducting ablation studies~\cite{warstadt2022artificial}.
In this study, we follow this spirit and investigate the role of trials and demonstrations through systematic computational experiments, assessing its role in neural word acquisition.

\vspace*{-5pt}
\subsection{Interaction in human language learning}

Social interactions are crucial in language acquisition, and the role of caregivers' feedback has been extensively explored in the field of developmental psychology~\citep{bates1975acquisition,saxton2005negative,warlaumont2014social}.
While our findings suggest that corrective feedback could enhance efficient neural word learning, these results should not be generalized to human language learning.
We discuss more about the role of interaction in human language learning in Appendix~\ref{sec:human-app}.

%% file: sec/06-conclusion.tex
\vspace*{-5pt}
\section{Conclusion}
\vspace*{-5pt}
\label{sec:conclusion}

This research introduces a trial-and-demonstration (TnD) learning framework to examine the effectiveness of corrective feedback in neural word acquisition through systematically controlled experiments, assessing how the interplay between student trials and teacher demonstrations contributes to learning efficiency in neural language models. 
We find that (1) TnD learning accelerates neural word acquisition across student models of different sizes; (2) the teacher's choices of words influence students' word-specific learning efficiency; and (3) a practice-makes-perfect effect is evident by a strong correlation between the frequency of words in trials and their respective learning curves.
Our findings confirm the crucial role of interaction in efficient word learning with language models.

\vspace*{-8pt}
\paragraph{Additional experiments.}

Due to the limited space and primary research scope of this work, we present experiments about other possible baselines besides controllable ablation, downstream natural language understanding task performances, and the robustness of the findings under different hyper-parameters in Appendix~\ref{sec:experiment-appendix}.

\section*{Limitations}
\label{sec:limitations}

\paragraph{Iterative setting.}
This experiment can be conducted iteratively by replacing the teacher model with the student model from previous iterations. 
While this iterative approach is intriguing, it introduces new complexities that require significant modifications to the current controlled ablation studies. 
We defer exploration of this approach to future work, as the current study focuses on examining the roles of trials and demonstrations.

\vspace*{-8pt}
\paragraph{The reward.}
Our study is limited by the use of a single reward model focused on corrective feedback. 
More realistic scenarios should also encompass communicative feedback, with the success of communication serving as a reward. 
Additionally, \citet{thorndike1911animal} proposed the idea that a child might instinctively feel satisfaction from producing a sound that echoes a meaningful memory. 
The design of such an intrinsic reward model is cognitively intriguing, and could aid in scaling student models without an external reward model, offering potential benefits for engineering applications. 

\vspace*{-8pt}
\paragraph{The reward model.}
We employ a robust language model (LLaMA-2-7B) as a reward model to concentrate on the roles of trials and demonstrations without concerns about reward quality. 
Future research should explore the impact of using less accurate reward models.

\vspace*{-8pt}
\paragraph{The tokenizer.}
One limitation of our approach is the reliance on the Byte Pair Encoding (BPE) tokenizer~\citep{sennrich2016neural} inherited from GPT-2. 
Ideally, the method should be tokenizer-free to facilitate the learning of early language elements such as sound effects and animal sounds, e.g., ``baa-baa'' in CDI~\citep{fenson2007macarthur}, which can be crucial for a more natural and foundational language acquisition process.

\vspace*{-8pt}
\paragraph{Other languages.}
The present study focuses on English as the subject of investigation due to the available corpus resources. 
Future research should consider exploring other languages.

%% file: sec/07-acknowledge.tex
\section*{Acknowledgement}

This work was supported in part by NSF IIS-1949634 and NSF SES-2128623.
Ziqiao Ma is supported in part by the Weinberg Cognitive Science Fellowship.
The authors extend their special appreciation to Susan Gelman and Freda Shi for their valuable feedback.
The authors would like to thank Run Peng, Yichi Zhang, Jacob Sansom, Zheyuan Zhang, Xuejun Zhang for proofreading and their helpful input during discussions.
We thank all anonymous reviewers for their feedback.

%% file: sec/a1-method.tex
\section{Trials-and-Demonstrations (TnD) Details and Discussions}
\label{sec:method-appendix}

\vspace*{-5pt}
\subsection{Interaction in human language learning}
\label{sec:human-app}

To unveil the role of social interactions in language acquisition, a prominent body of research has focused on \textit{pragmatic inference} -- that children exhibit the ability to refine their linguistic knowledge by inferring the communicative intents of others~\citep{senju2008gaze,yurovsky2017beyond,bohn2019pervasive}.
In addition, other researchers have argued for the importance of caregivers' \textit{feedback} to the language produced by children~\citep{warlaumont2014social}, in the form of descriptions, explanations, corrections, etc, to human language development.
For example, extensive efforts have been made to examine the effects of \textit{communicative feedback} on language acquisition, both in developmental psychology~\citep{bates1975acquisition,snow1996learning,nikolaus2023communicative} and computational modeling~\citep{nikolaus2021modeling,liu2022computational}. 
This type of feedback emphasizes the explicit negotiation of mutual understanding with the conversational partner to achieve and maintain common ground~\cite{clark1996using}.
While communicative feedback emphasizes the success and failure of communication, the feedback we study in this study is more akin to \textit{corrective feedback}.
This type of feedback involves responses from caregivers, which offer corrections to possible errors in children's speech, including variants such as negative evidence, reformulations, or recasts~\citep{farrar1992negative,saxton2000negative,chouinard2003adult,saxton2005negative,hiller2016corrective}.
Although corrective feedback is shown to be helpful in second language acquisition~\citep{el2002corrective,ellis2006implicit,bitchener2011written}, researchers largely dispute its availability and effectiveness in human first language acquisition~\citep{brown1970derivational,marcus1993negative}.
While our findings suggest that corrective feedback through demonstrations can enhance efficient neural language learning, these results should not be generalized to human language learning, where demonstrations are much less frequent in the noisy feedback that children typically receive.

\vspace*{-5pt}
\subsection{Relationship with RLHF}
\label{sec:rlhf-app}

Reinforcement learning enables language systems to learn from feedback in the form of rewards from games~\citep{narasimhan2015language,he2016deep} or heuristic scores~\citep{ranzato2016sequence,nikolaus2021modeling}.
Recent advancements in reinforcement learning from human feedback (RLHF) have generated considerable excitement, especially in its application to large language models such as ChatGPT~\citep{chatgpt2022openai}. 
Reinforcement learning is employed to align the model's policy with human preferences, utilizing human-annotated preference data~\citep{ziegler2019fine,ouyang2022training} or AI models acting as proxies for human judgment~\citep{bai2022constitutional,lee2023rlaif}.
We refer to \citet{kaufmann2023survey} and \citet{zheng2023secrets} for more details.
Our work leverages the implementation infrastructure of RLHF but diverges in significant ways: whereas RLHF aims to align an existing language model with human preferences, our objective is to ``babysit'' a language model from scratch using reinforcement learning, specifically to model the process of receiving and integrating corrective feedback.

\vspace*{-5pt}
\subsection{Preliminaries}
\label{sec:rl-app}

Inspired by recent work in RLHF, we use Proximal Policy Optimization (PPO) \cite{schulman2017proximal} to train the student model. 
Consider a model \(\pi_{\theta}\) whose the current state \(s_i\) is a sequence of tokens \(s_i={t_1,\cdots,t_i}\), and receives a reward \(r_i\).
We outline the key components and refer to \citet{zheng2023secrets} for more details.

\paragraph{Clipped surrogate objective.}
The clipped surrogate objective is defined as:
\vspace*{-4.5pt}
\begin{equation}
\mathcal{L}^{\textrm{pg}}_\theta = \mathbb{E}_i\big[\min \big(\Pi_i(\theta), \textsc{clip}\big(\Pi_i(\theta),\epsilon\big)\big) A_i\big]
\vspace*{-5pt}
\end{equation}
where
\vspace*{-5pt}
\begin{equation}
\Pi_i(\theta) := \frac{\pi_{\theta}(t_{i+1}\mid s_i)}{\pi_{\theta_{\textrm{old}}}(t_{i+1}\mid s_i)}
\vspace*{-5pt}
\end{equation}
The \textsc{clip} function clips the value within \((1-\epsilon,1+\epsilon)\), which regularizes the policy from drastic changes to ensure robustness.

\paragraph{Generalized advantage estimation.}
The advantage function \(A_i\) at step $i$ is estimated by the Generalized Advantage Estimation (GAE) algorithm \cite{schulman2015high} for a balanced bias-variance trade-off:
\vspace*{-4.5pt}
\begin{equation}
A_i = \sum_{k=0}^{\infty} (\gamma \lambda)^k \delta_{i+k}
\vspace*{-5pt}
\end{equation}
where \(\gamma\) is the discount factor, \(\lambda\) is a hyperparameter controlling the trade-off between bias and variance, and \(\delta_{i} = r_{i} + \gamma V(s_{i+1}) - V(s_{i})\) is the temporal difference (TD) error. The reward $r$ is defined in \Cref{sec:reward}.

\paragraph{Proximal policy optimization.}
During the same time, PPO also estimates and optimizes its value function \(V_{\theta_{vh}}\) with MSE loss:
\vspace*{-5pt}
\begin{equation}
\mathcal{L}^{\text{value}}_{\theta_{vh}} = \mathbb{E}_i\left[\left(V_{\theta_{vh}}(s_i) - \hat{V_i}\right)^2\right]
\vspace*{-4pt}
\end{equation}
where \(V_{\theta_{vh}}(s_i)\) is the estimated value, and \(\hat{V_i}\) is the target value from GAE.
Differing from conventional implementation, our approach didn't employ a reference model for the KL penalty. This deviation emphasizes the language model's evolutionary nature during its training, enabling more significant updates and eliminating biases from over-adherence to a reference model. 
The final reinforcement learning loss is a linear combination:

\vspace*{-10pt}
\begin{equation}
\label{eq:ppo}
    \mathcal{L}^{\textrm{ppo}}_\theta = \mathcal{L}^{\text{pg}}(\theta) + c \cdot \mathcal{L}^{\text{value}}_{\theta_{vh}}
\vspace*{-3pt}
\end{equation}
where \(c\in[0,1]\).

%% file: sec/a2-experiment.tex
\section{Additional Experiments, Results, and Discussions}
\label{sec:experiment-appendix}

\subsection{Reproducibility}
\label{sec:reproduce}

\paragraph{Test vocabulary.}
We include the list of words and the evaluation datasets within the code.\footnote{Our code and data will be made publicly available upon the acceptance of this work.}

\paragraph{Training details.}
We randomly initialize the GPT-2 student models.
For each combination of corpus and baseline, the training process is conducted 5 times for 10k steps, each with a different random seed. 
We utilize the top-$k$ decoding strategy for language generation, setting $k=20$.
The learning rate for reinforcement learning is set to \(2e^{-5}\). 
We follow the default setting for other PPO hyper-parameters, such as a clip range (\(\epsilon=0.2\)) from the \href{https://github.com/huggingface/trl/}{TRL} library.
All other hyper-parameters for causal language modeling, including the learning rate set at \(1e^{-4}\) and a batch size of 128, remain consistent with the training setup of GPT-2 by~\citet{chang2022word}.

\paragraph{Checkpointing.}
We save the intermediate steps: [2, 4, 6, 8, 10, 12, 14, 16, 18, 20, 25, 30, 35, 40, 45, 50, 55, 60, 65, 70, 75, 80, 85, 90, 95, 100, 110, 120, 130, 140, 150, 160, 170, 180, 190, 200, 210, 220, 230, 240, 250, 260, 270, 280, 290, 300, 310, 320, 330, 340, 350, 360, 370, 380, 390, 400, 410, 420, 430, 440, 450, 460, 470, 480, 490, 500, 550, 600, 650, 700, 750, 800, 850, 900, 950, 1000, 1500, 2000, 2500, 3000, 3500, 4000, 4500, 5000, 5500, 6000, 6500, 7000, 7500, 8000, 8500, 9000, 9500, 10000, 20000, 30000, 40000, 50000, 60000, 70000, 80000, 90000, 100000].

\paragraph{Computational resources.}
In \texttt{TnD} training, each experiment to train a student model using \texttt{TnD} is conducted on 2 A40 GPUs for 36 hours.
In teacher model pre-training, we distribute the computation over 4 A40 GPUs with batch size 32 per device for 20 hours.
To fine-tune the neural age predictor, we use a LLaMA-2-7B model~\citep{touvron2023llama} with regression head on our developmental trajectory (text and step pairs) dataset. 
To ensure the prediction quality and save computation resources, we fine-tune our model using mixed precision on all parameters. 
The training is distributed over 8 A40 GPUs with batch size 8 per device with fully-shared data parallel. 
The learning rate is \(5e^{-5}\).

\input{sec/input/trial-appendix}

\vspace*{-5pt}
\subsection{Additional Results}
\label{sec:more-trial-examples}

\paragraph{nAoA distributions.}

We present the ridgeline and scatter plot of words and their neural age of acquisition (nAoA) in BabyLM Corpus (Figure~\ref{fig:scatter-babylm}) and BookCorpus (Figure~\ref{fig:scatter-bkps}).

\vspace*{-5pt}
\paragraph{Surprisal and learning curves.}

\input{sec/input/sigmoid_plateau}
We observe a similar pattern reported by \citet{chang2022word} that learning curves tend to level off at a local plateau, which aligns with the unigram surprisal.
This phenomenon renders the single-sigmoid model unreliable for capturing the complexity, and we adopt a double-sigmoid function to fit the learning curve.
We run regression between the plateau of double-sigmoid curves and the unigram surprisals calculated from all sources of word occurrences (\Cref{fig:sigmoid-combined}). 
We find a strong correspondence between the plateau and unigram frequency, suggesting that the double-sigmoid function is a better option than the single-sigmoid function to fit learning curves.
To analyze longer learning curves of up to 1M steps, more complex functions such as linear GAMs have been adopted~\citep{chang2023characterizing}.
For our purposes, a double-sigmoid function suffices.

\vspace*{-5pt}
\subsection{Teacher's word preferences in demonstrations affect student}
\label{sec:teacher}

To explore how the teacher model's word selection impacts students' language development, we repeat the experiments where a chosen set of 40 words for each test vocabulary is excluded from teacher demonstrations. 
To ensure fluent generation, we maintain the presence of essential functional words so these words don't appear in the 40 chosen words.
During the language generation process by the teacher model, if a word from this set is to be decoded, we select the next best alternative, ensuring these words were never presented in teacher demonstrations.
We depict the learning curves for these excluded words in Figure~\ref{fig:mask} and present the nAoA in Table~\ref{tab:aoa-masked}.
Our findings indicate that the teacher model's word choices significantly influence the efficiency of word acquisition by the student model. 
The absence of words from teacher demonstrations leads to slower learning speed for the student models, as evidenced by a higher \texttt{nAoA}, although the student models are ultimately able to learn these words from the corpus and their trials.

\input{sec/input/superglue}

\vspace*{-5pt}
\paragraph{Additional per-word learning curves.}
We report additional examples of per-word learning curves and cumulative word frequency over time in \Cref{fig:more-trial}. 
Predicates and function words, such as ``back'' and ``go'', have a high correlation between surprisal and cumulative word frequency on the student model's trial, shown in \Cref{fig:more-trial}(a-f). Nouns such as ``light'' and ``car'', however, depict a less correlation between the word's learning curve and trial's frequency, shown in \Cref{fig:more-trial}(g-h).

\vspace*{-5pt}
\subsection{Downstream evaluation on NLU tasks}
\label{sec:nlu}

We evaluate the final model on downstream natural language understanding (NLU) tasks.
Specifically, we fine-tune the final CLM and TnD models (from the BabyLM corpus) on the BabyLM round 1 NLU evaluation set, which is based on (Super)GLUE~\citep{wang2018glue,wang2019superglue}. 
Table~\ref{tab:super-glue} shows that the TnD model performs on par with the CLM model, slightly better overall. 
We observe that the TnD model did significantly better on the Recognizing Textual Entailment (RTE, which requires determining inferential relationships between hypothesis and premise) task, but underperforms on the Question-Answering NLI (QNLI, which requires comprehending longer paragraphs).

\vspace*{-5pt}
\subsection{Discussion and experiment on other possible baselines}
\label{sec:baselines}

\vspace*{-3pt}
\paragraph{Causal language modeling using teacher-generated text (naive content distillation).}
One possible baseline is to re-run the \texttt{CLM} baseline using teacher/student-generated texts, rather than the corpus text, at a non-interactive step.
This experiment is controlled over the language input to the \texttt{CLM} and \texttt{TnD} baselines, but is not a fair setting for \texttt{CLM} as the student's trials are usually of poor quality in initial steps.
We replace the PPO updates in \texttt{TnD} with causal language modeling on both texts from the student model's trial and the teacher model's demonstration. 
We find that the resulting \texttt{CLM}$_\texttt{TnD}$ baseline achieves an almost identical overall learning curve as \texttt{CLM} (\Cref{fig:double-clm-step}).

\vspace*{-3pt}
\paragraph{Causal language modeling using teacher-generated text (naive content distillation).}
Our main difference from the existing knowledge distillation methods for LLMs is that we train a student model from scratch, rather than requiring a pre-trained but weaker student model. 
For example, \citet{ko2024distillm} adopt the T5 language model as the student. 
Closest to our work is \citet{nikolaus2021modeling}, which also adopts an ablation study setup. 
We did not compare to their method as a baseline as they noticed that combining production and perception-based language learning does not work from scratch. 
To the best of our knowledge, our TnD method is the only interactive language learning algorithm that functions well with a student model from scratch.

\vspace*{-5pt}
\paragraph{Using the corpus sentences as demonstrations.}
While it is possible to directly use the ground truth sentences from the training corpus as demonstrations~\citep{ter2022towards}, it can be very difficult to adapt the teacher model's behaviors for our controlled studies.
As a result, we use a pre-trained language model as the proxy for demonstration generation, rather than using the original text.
In our preliminary experiments, using ground truth sentences from the training corpus as demonstrations do not lead to a noticeable difference from model-generated demonstrations.

\subsection{The robustness of results and findings over hyper-parameters}
\label{sec:parameters}

\input{sec/input/different-clm-combined}
\input{sec/input/param-combined}

\vspace*{-5pt}
\paragraph{Learning rate}
To evaluate the robustness of results over different PPO learning rates, we run our experiment with learning rate = \(8e^{-6},1e^{-5},2e^{-5},3e^{-5}\) (Figure~\ref{fig:lr}). 
We find that a smaller learning rate results in a later age of acquisition compared to a higher learning rate. 
Nevertheless, our findings remain robust across different learning rates, although a higher learning rate can lead to unstable training and poorer end performance.

\vspace*{-5pt}
\paragraph{Alternating frequency}
We run our experiment on different alternating frequencies to study its impact on the \texttt{TnD} framework. To perform a systematic study, we group the alternating schedule \([c,r]\) by their PPO/CLM steps ratio. 
We experiment with the settings when the ratio equals to \(1\) (\([1,1]\), \([2,2]\)), \(2\) (\([2,1]\), \([4,2]\)), \(3\) (\([3,1]\), \([6,2]\)), and \(4\) (\([4,1]\)), as presented in Figure~\ref{fig:ratio}. 
We find that the alternating frequencies under the same ratio lead to similar performance.
Our findings are robust over different alternating frequencies, except the ratio \(= 1\) leading to unstable training and poor end performance.

%% file: sec/input/trial-appendix.tex
\begin{figure*}[!h]
    \centering
    \begin{subfigure}[t]{.23\textwidth}
        \centering
        \includegraphics[width=1.1\linewidth]{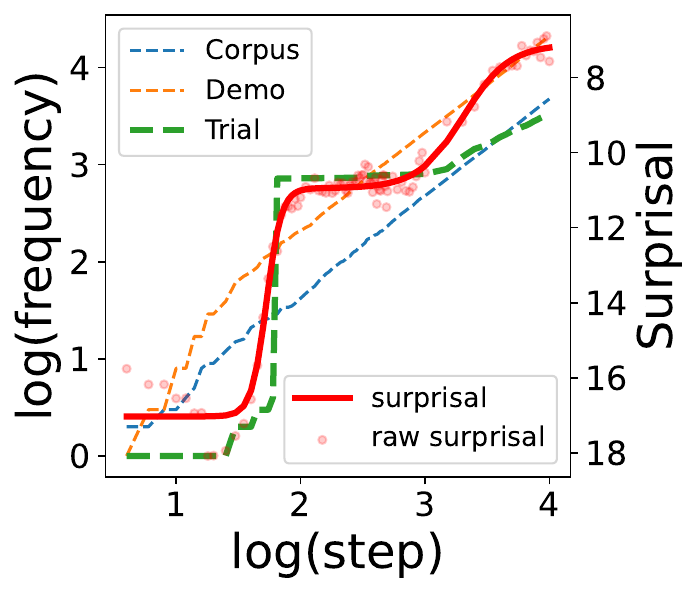}
        \vspace*{-20pt}
        \caption{The word ``hear.''}
    \end{subfigure}
    ~
    \begin{subfigure}[t]{.23\textwidth}
        \centering
        \includegraphics[width=1.1\linewidth]{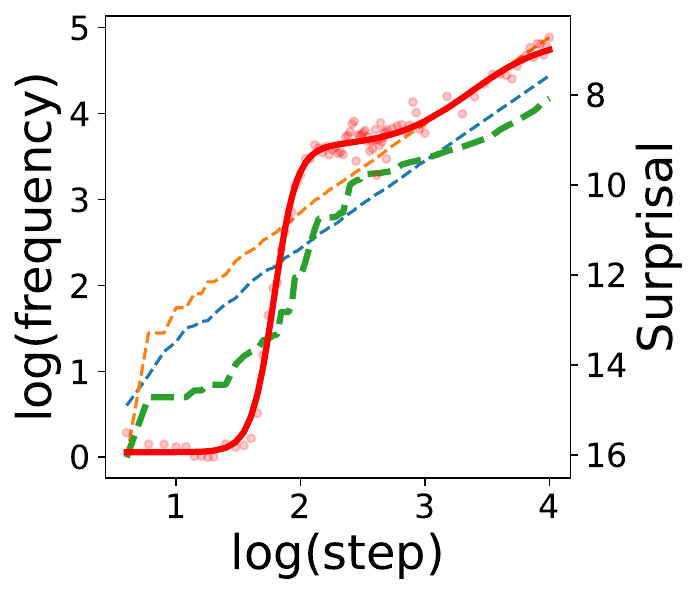}
        \vspace*{-20pt}
        \caption{The word ``there.''}
    \end{subfigure}
    ~
    \begin{subfigure}[t]{.23\textwidth}
        \centering
        \includegraphics[width=1.1\linewidth]{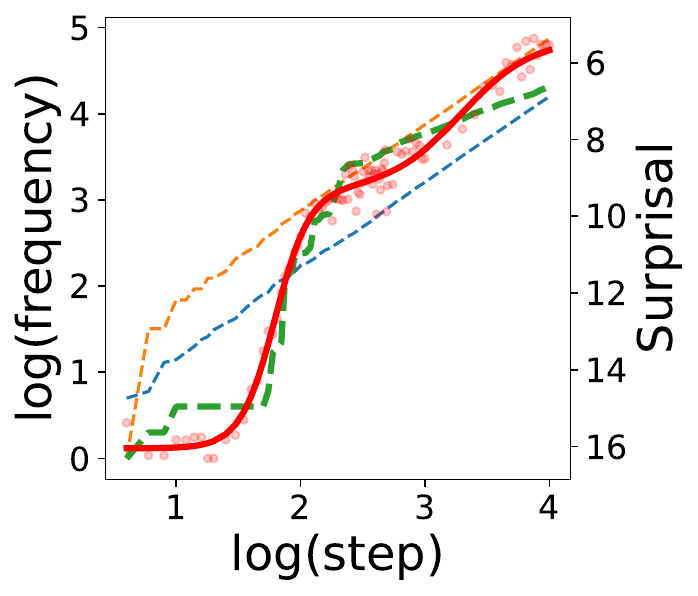}
        \vspace*{-20pt}
        \caption{The word ``go.''}
    \end{subfigure}
    ~
    \begin{subfigure}[t]{.23\textwidth}
        \centering
        \includegraphics[width=1.1\linewidth]{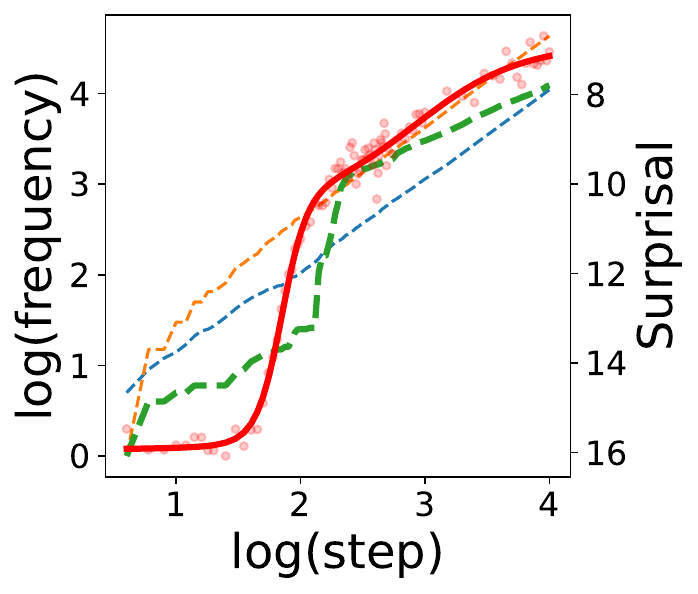}
        \vspace*{-20pt}
        \caption{The word ``take.''}
    \end{subfigure}
    ~
    \begin{subfigure}[t]{.23\textwidth}
        \centering
        \includegraphics[width=1.1\linewidth]{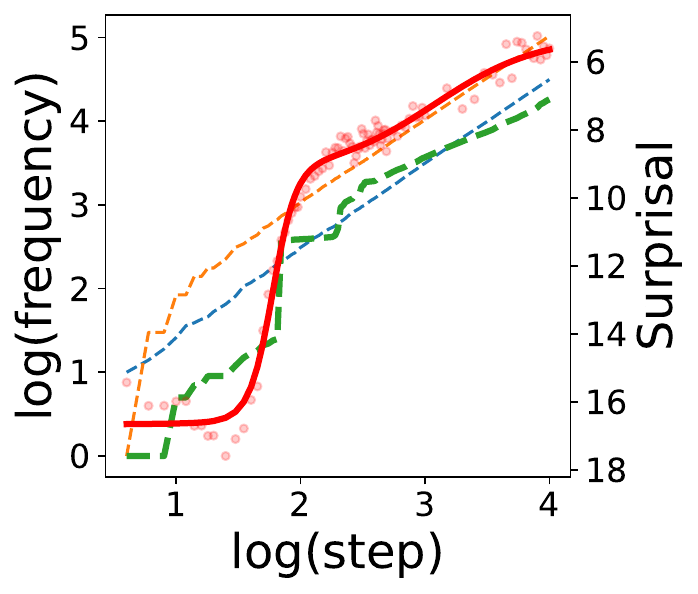}
        \vspace*{-20pt}
        \caption{The word ``if.''}
    \end{subfigure}
     ~
    \begin{subfigure}[t]{.23\textwidth}
        \centering
        \includegraphics[width=1.1\linewidth]{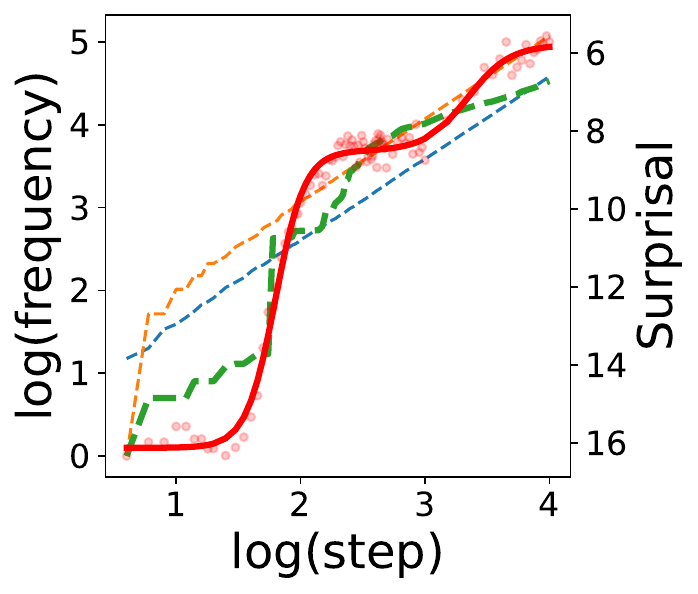}
        \vspace*{-20pt}
        \caption{The word ``back.''}
    \end{subfigure}    
    ~
    \begin{subfigure}[t]{.23\textwidth}
        \centering
        \includegraphics[width=1.1\linewidth]{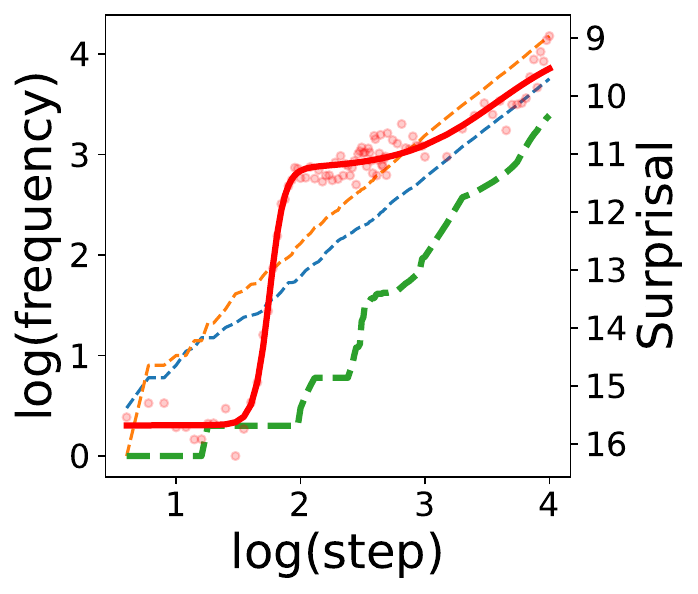}
        \vspace*{-20pt}
        \caption{The word ``light.''}
    \end{subfigure}
    ~
    \begin{subfigure}[t]{.23\textwidth}
        \centering
        \includegraphics[width=1.1\linewidth]{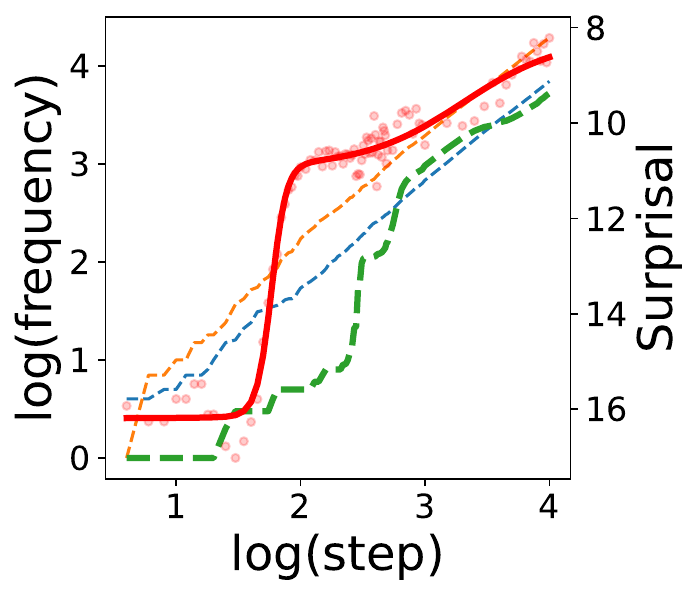}
        \vspace*{-20pt}
        \caption{The word ``car.''}
    \end{subfigure}
    \vspace*{-10pt}
    \caption{Examples of per-word learning curves and cumulative word frequency in BookCorpus. \vspace*{-10pt}}
    \label{fig:more-trial}
\end{figure*}

%% file: sec/input/sigmoid_plateau.tex

\begin{figure}[!t]
    \centering
    \begin{subfigure}[t]{.23\textwidth}
        \centering
        \includegraphics[width=1.05\linewidth]{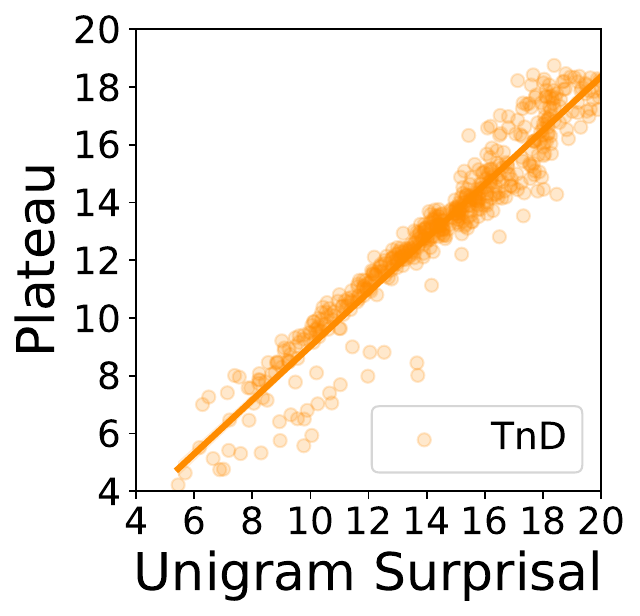}
        \vspace*{-20pt}
        \caption{Plateau v.s. unigram surprisal in BookCorpus.}
        \label{fig:sigmoid-bkps}
    \end{subfigure}
    ~
    \begin{subfigure}[t]{.23\textwidth}
        \centering
        \includegraphics[width=1.05\linewidth]{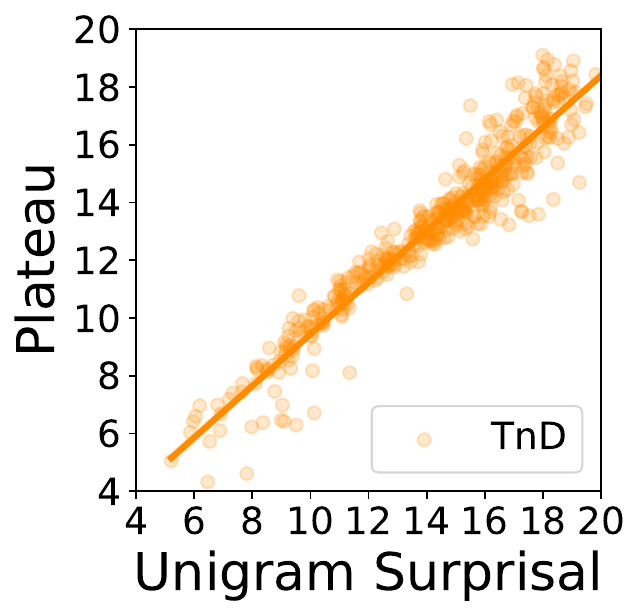}
        \vspace*{-20pt}
        \caption{Plateau v.s. unigram surprisal in BabyLM.}
        \label{fig:sigmoid-babylm}
    \end{subfigure}   
    \vspace*{-5pt}
    \caption{Sigmoid plateau v.s. unigram surprisal. \vspace{-15pt}}
    \label{fig:sigmoid-combined}
\end{figure}

%% file: sec/input/superglue.tex
\begin{table*}[!t]
\centering
\scalebox{0.9}{
    \begingroup
    \setlength{\tabcolsep}{2pt}
    \renewcommand{\arraystretch}{0.8}
    \begin{tabular}{lcccccccccccc}
    \toprule
    Model & Average & CoLA & SST-2 & MRPC & QQP & MNLI & MNLI$_\textrm{mm}$ & QNLI & RTE & BoolQ & MultiRC & WSC \\
    \cmidrule(r){1-1} \cmidrule(r){2-2} \cmidrule(r){3-13}
    \texttt{CLM} + BabyLM  & 67.2 & \textbf{69.8} & \textbf{84.4} & \textbf{76.2} & \textbf{79.1} & 67.2 & 68.0 & \textbf{68.5} & 48.5 & 63.6 & 52.0 & 61.4 \\
    \texttt{TnD} + BabyLM & \textbf{67.5} & 66.6 & 82.5 & 75.4 & 79.0 & \textbf{67.4} & \textbf{69.0} & 61.0 & \textbf{60.6} & \textbf{64.9} & \textbf{54.5} & \textbf{61.4} \\
    \cmidrule(r){1-1} \cmidrule(r){2-2} \cmidrule(r){3-13}
    \texttt{CLM} + BookCorpus  & 65.5 & \textbf{67.6} & \textbf{85.4} & 73.2 & \textbf{78.3} & 66.0 & 67.1 & 60.1 & 45.5 & 65.3 & 51.0 & \textbf{61.4} \\
    \texttt{TnD} + BookCorpus & \textbf{65.8} & 67.5 & 83.3 & \textbf{77.3} & 77.5 & \textbf{66.7} & \textbf{67.4} & \textbf{66.4} & \textbf{45.5} & \textbf{67.5} & \textbf{51.2} & 53.0 \\
    \bottomrule
    \end{tabular}
\endgroup}
 \vspace{-5pt}
\caption{We present the NLU evaluation results based on the BabyLM Challenge, which consists of tasks selected from (Super)GLUE. We fine-tune the final models developed on both BookCorpus and BabyLM datasets. \vspace{-10pt}}
\label{tab:super-glue}
\end{table*}

%% file: sec/input/different-clm-combined.tex

\begin{figure}[!t]
    \vspace{-10pt}
    \centering
    \begin{subfigure}[t]{.23\textwidth}
        \centering
        \includegraphics[width=1.05\linewidth]{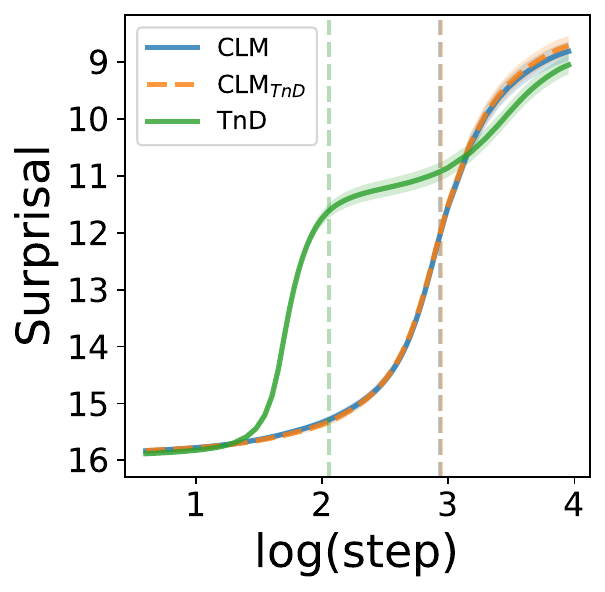}
        \vspace*{-20pt}
        \caption{Overall surprisal v.s. training step on BabyLM.}
        \label{fig:double-clm-freq}
    \end{subfigure}
    ~
    \begin{subfigure}[t]{.23\textwidth}
        \centering
        \includegraphics[width=1.05\linewidth]{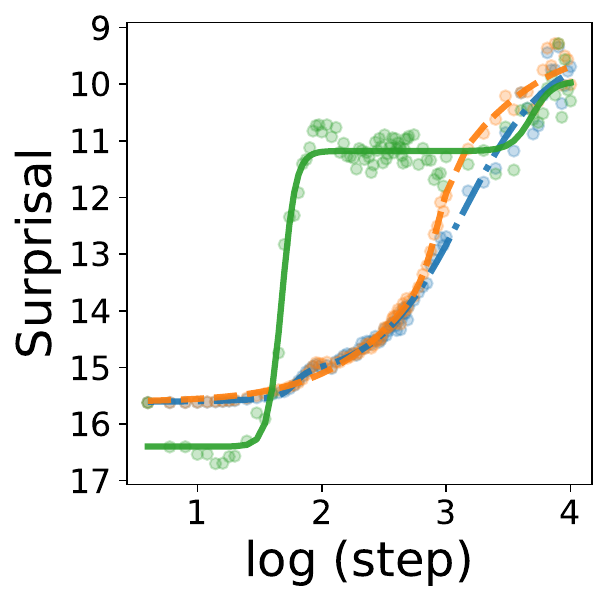}
        \vspace*{-20pt}
        \caption{Surprisal v.s. training step on word ``fine''.}
        \label{fig:double-clm-step}
    \end{subfigure}   
    \vspace*{-5pt}
    \caption{Comparison between CLM on model predicted text, CLM on train corpus, and TnD. Evaluated on \texttt{CDI} words on BabyLM. \vspace*{-15pt}}
    \label{fig:distill}
\end{figure}

%% file: sec/input/param-combined.tex

\begin{figure}[!t]
    \vspace{-10pt}
    \centering
    \begin{subfigure}[t]{.23\textwidth}
        \centering
        \includegraphics[width=1.05\linewidth]{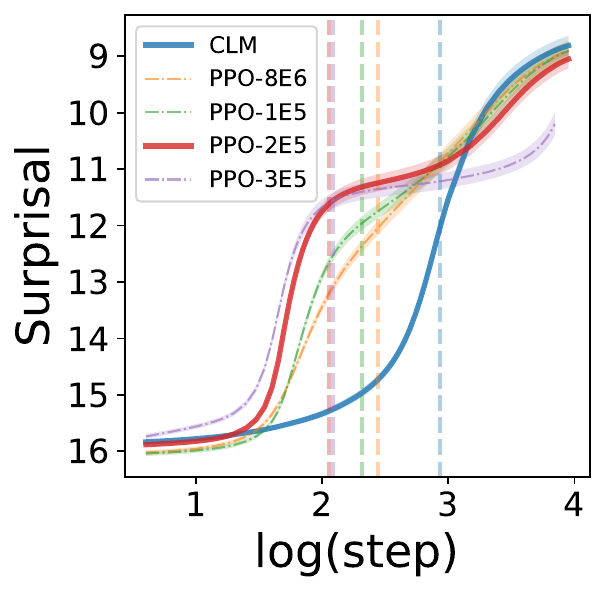}
        \caption{Effect of PPO learning rate on \texttt{CDI} words on BabyLM.}
        \label{fig:lr}
    \end{subfigure}
    ~
    \begin{subfigure}[t]{.23\textwidth}
        \centering
        \includegraphics[width=1.05\linewidth]{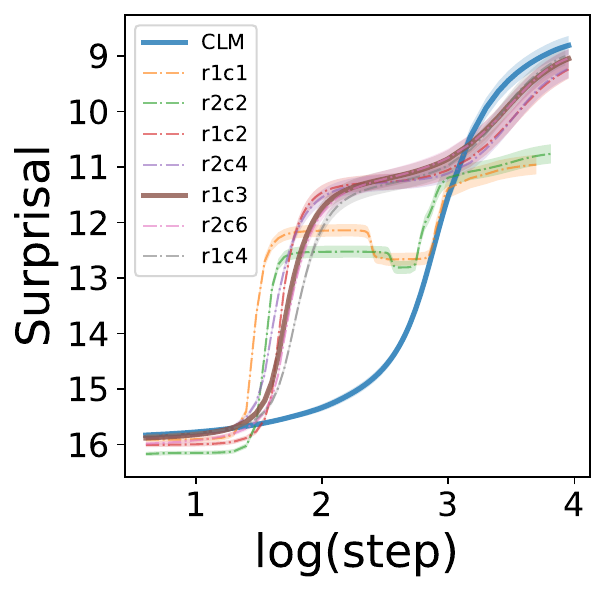}
        \caption{Effect of alternating frequency on \texttt{CDI} words on BabyLM.}
        \label{fig:ratio}
    \end{subfigure}   
    \vspace*{-5pt}
    \caption{The robustness of the learning curve results over other hyper-parameters, e.g., learning rates and alternating schedules. \vspace{-15pt}}
    \label{fig:params}
\end{figure}

%% file: sec/a3-ethics.tex
\section{Limitations, Licenses, and Risks}
\label{sec:risks-appendix}

\subsection{Artifacts and licenses}

Our work largely relies on publicly available datasets such as BookCorpus~\cite{zhu2015aligning} and BabyLM~\cite{warstadt2023findings}, and pre-trained models such as GPT-2~\cite{radford2019language} and LLaMA-2~\cite{touvron2023llama}.
We strictly follow the LLaMA license and limit the scope of the LLaMA model to academic research only. 
We report a list of licenses for all datasets and models used in our experiment in Table~\ref{tab:license}.
\input{sec/input/license}

\vspace*{-5pt}
\subsection{Ethical concerns and risks}

This work does not depend on human annotators or human subjects for interactive experiments.
We leverage open datasets and model-generated content for training that could contain biases and sensitive contents inherited, which may cause fairness issues in the final model when applied to practical applications.
Future research should be done to look into these issues, potentially by designing fairness-aware reward models.

\input{sec/input/scatter}

%% file: sec/input/license.tex

\begin{table}[!t]
\centering
    \scalebox{0.9}{
    \begingroup
    \setlength{\tabcolsep}{2pt}
    \renewcommand{\arraystretch}{0.8}
    \begin{tabular}{@{}lll@{}}
    \toprule
    Dataset & URL & License
    \\ \midrule
    BookCorpus & \href{https://github.com/soskek/bookcorpus/blob/master/LICENSE}{Link} & MIT License\\
    BabyLM & \href{https://github.com/babylm/evaluation-pipeline?tab=MIT-1-ov-file}{Link} & MIT License
    \\ \midrule \midrule
    Models & URL & License 
    \\ \midrule
    GPT-2 & \href{https://github.com/openai/gpt-2/blob/master/LICENSE}{Link} & MIT License\\
    LLaMA-2 & \href{https://ai.meta.com/llama/license/}{Link} & Llama License\\
    \bottomrule
    \end{tabular}
    \endgroup}
    \vspace*{-5pt}
    \caption{License information for the code base in our experiment.\vspace*{-20pt}}
    \label{tab:license}
\end{table}

%% file: sec/input/scatter.tex
\begin{figure*}[!h]
    \centering
    \vspace*{-16pt}
    \hspace*{9mm}
    \vspace{-33.5pt} 
    \includegraphics[width=0.75\linewidth]{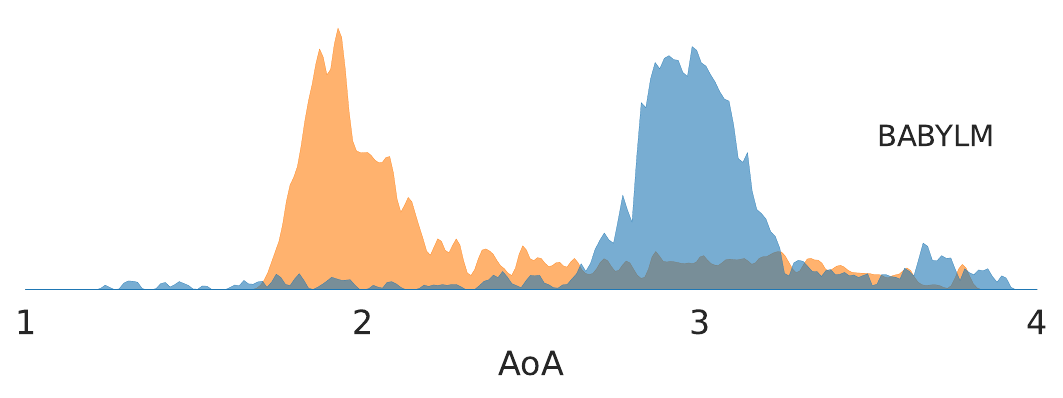}
    \includegraphics[width=0.82\linewidth]{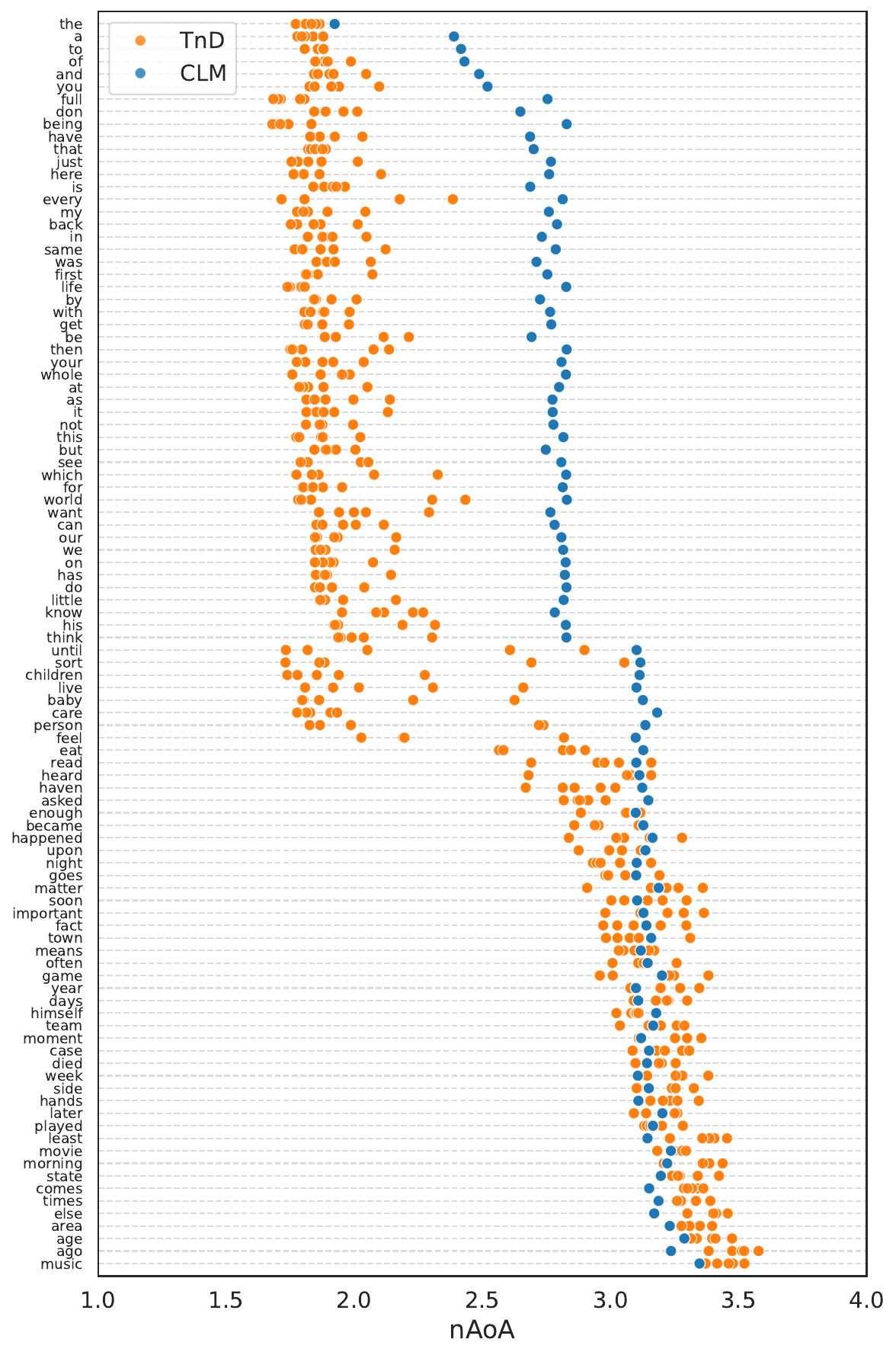}
    \vspace*{-10pt}
    \caption{The ridgeline and scatter plot of words and their neural age of acquisition (nAoA) in BabyLM Corpus.}
    \label{fig:scatter-babylm}
\end{figure*}

\begin{figure*}[!h]
    \centering
    \vspace*{-16pt}
    \hspace*{9mm}
    \vspace{-33.5pt} 
    \includegraphics[width=0.75\linewidth]{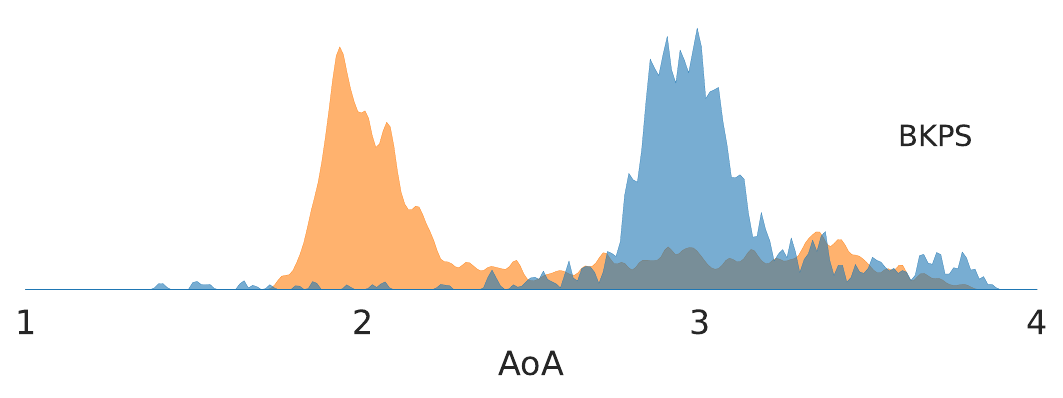}
    \includegraphics[width=0.82\linewidth]{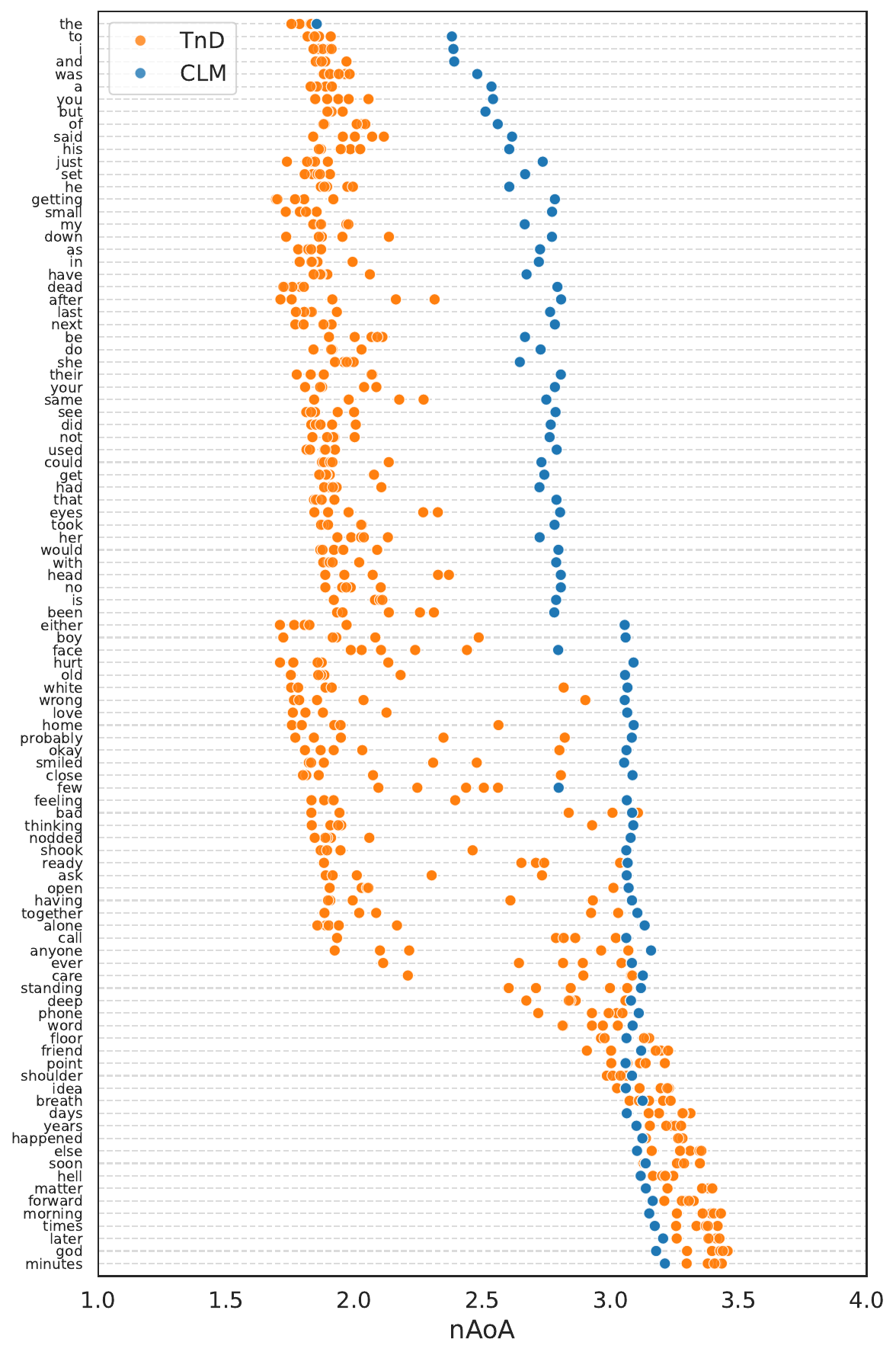}
    \vspace*{-10pt}
    \caption{The ridgeline and scatter plot of words and their neural age of acquisition (nAoA) in BookCorpus.}
    \label{fig:scatter-bkps}
\end{figure*}

